\documentclass{llncs}

\usepackage{layout}
\usepackage[T1]{fontenc}
\usepackage[utf8]{inputenc}
\usepackage{graphicx}
\usepackage{amsmath}
\usepackage[french]{babel}

\begin{document}

\title{Un duel probabiliste pour départager\\ deux présidents (LIA @ DEFT'2005)}

\author{Marc El-Bèze\inst{1} \and Juan-Manuel Torres-Moreno\inst{1,}\inst{2} \and Frédéric Béchet\inst{3}}

\authorrunning{El-Bèze et al.}

\institute{Laboratoire Informatique d'Avignon\\
Université d'Avignon \\
BP 91228 84911 Avignon Cedex 09, France\\
\and
 Polytechnique Montréal\\
 CP 6079 Succ. centre-Ville, H3C 3A7 Montréal (Québec) Canada\\
 \and
 Laboratoire Informatique et Systèmes - LIS UMR7020, \\
 Aix-Marseille Université France\\
\email{juan-manuel.torres@univ-avignon.fr}
}

\date{\today}

\maketitle

\begin{abstract} 

\noindent We present a set of probabilistic models applied to binary classification as defined in the DEFT'05 challenge. The challenge consisted a mixture of two differents problems in Natural Language Processing : identification of author (a sequence of François Mitterrand's sentences might have been inserted into a speech of Jacques Chirac) and thematic break detection (the subjects addressed by the two authors are supposed to be different). Markov chains, Bayes models and an adaptative process have been used to identify the paternity of these sequences. A probabilistic model of the internal coherence of speeches which has been employed to identify thematic breaks. Adding this model has shown to improve the quality results. A comparison with different approaches demostrates the superiority of a strategy tha combines learning, coherence and adaptation. Applied to the DEFT'05 data test the results in terms of precision (0.890), recall (0.955) and {\itshape Fscore} (0.925) measure are very promising.

\end{abstract}

\section{Introduction}

Dans le cadre des conférences TALN\footnote{Traitement Automatique des Langues Naturelles.} et RECITAL\footnote{Rencontre des Etudiants Chercheurs en Informatique pour le Traitement Automatique des Langues.} tenues en juin 2005 à Dourdan (Fran\-ce), un atelier a été organisé autour du défi de fouille textuelle proposé par \cite{aze:2005}. Ce défi portait le nom de DEFT'05 (DÉfi Fouille de Textes). Il a été motivé par le besoin de mettre en place des techniques de fouille de textes permettant soit d'identifier des phrases non pertinentes dans des textes, soit d'identifier des phrases particulièrement singulières dans des textes apparemment sans réel intérêt. Concrètement, il s'agissait de supprimer les phrases non pertinentes dans un corpus de discours politiques en français. Cette tâche est proche de la piste {\itshape Novelty} du challenge TREC \cite{soboro:2004} qui dans sa première partie consiste à identifier les phrases pertinentes puis, parmi celles-ci, les phrases nouvelles d'un corpus d'articles journalistiques. Pour mieux comprendre à quoi correspondait dans DEFT'05 la suppression des phrases non pertinentes d'un corpus de discours politiques \cite{alphonse:2005}, une brève description du but général\footnote{Le lecteur intéressé trouvera une description plus détaillée ainsi que les données et résultats dans le site officiel du défi : http://www.lri.fr/ia/fdt/DEFT05} s'impose. Un corpus de textes, allocutions officielles issues de la Présidence (1995-2005) de Jacques Chirac  a été fourni. Dans ce corpus, des passages issus d'un corpus d'allocutions (1981-1995) du Président François Mitterrand ont été insérés. Les passages d'allocutions de F. Mitterrand insérés sont composés d'au moins deux phrases successives et ils sont censés traiter une thématique différente\footnote{Par exemple, dans les allocutions de Jacques Chirac évoquant la politique internationale, les phrases de François Mitterrand introduites sont issues de discours traitant de politique nationale.}. Un corpus, formé de discours de J. Chirac entrecoupés d'extraits de ceux de F. Mitterrand, est ainsi constitué. 

Certaines informations sont supprimées de ce corpus afin de constituer les trois corpus ci-dessous :

\begin{itemize}
    \item Corpus C1 : aucune présence d'années ni de noms de personnes : ils ont été remplacés par les balises \textsc{<date>} et \textsc{<nom>} ;
    \item Corpus C2 : pas d'années : elles ont été remplacées par la balise \textsc{<date>} ;
    \item Corpus C3 : présence des années et des noms de personnes.
\end{itemize}

Le but du défi consistait à déterminer les phrases issues du corpus de F. Mitterrand introduites dans le corpus composé d'allocutions de J. Chirac. Ce but est commun aux trois tâches T1, T2 et T3 relatives aux trois corpus C1, C2 et C3 ont ainsi été définies. Intuitivement, la tâche T1 est la plus difficile des trois car le corpus afférent C1 contient moins d'informations que les deux autres. Les résultats (calculés uniquement sur les phrases de F. Mitterrand extraites) peuvent être évalués sur un corpus de test (T) avec des caractéristiques semblables à celui de développement (D) (cf. tableau \ref{tab:stats}), en calculant le {\itshape Fscore} :

\begin{equation}
Fscore(\beta)= \frac{(\beta^2+1)\times Pr\acute{e}cision \times Rappel}{\beta^2\times Pr\acute{e}cision + Rappel}
\end{equation}

Dans le cadre de DEFT'05, le calcul du {\itshape Fscore} retenu par les organisateurs a été effectué uniquement sur les phrases de Mitterrand, et il a été modifié \cite{alphonse:2005} comme suit (cette réécriture suppose évidemment que $\beta$ soit égal à 1 de façon à ne privilégier ni précision ni rappel) :

\begin{equation}
Fscore(\beta = 1)= \frac{2 \times \textrm{Nb\_phrases\_correctes\_extraites}}{\textrm{Nb\_total\_extraites} + \textrm{Nb\_total\_pertinentes}}
\end{equation}

\begin{itemize}
  \item {Nb\_phrases\_correctes\_extraites} : nombre de phrases qui appartiennent réellement au corpus de Mitterrand dans le fichier résultat fourni par le système ;
    \item { Nb\_total\_extraites} : nombre de phrases données dans le fichier résultat (que le système considère comme étant Mitterrand);
    \item   { Nb\_total\_pertinentes} : nombre total de phrases de Mitterrand dans le corpus de test.
\end{itemize}

\begin{table}[ht]
 \begin{center}
   \tabcolsep = 1.6\tabcolsep
   \begin{tabular}{rrrrr}
   \hline\hline
      Discours  & Phrases (D) & Mots (D) & Phrases (T) & Mots (T) \\
   \hline
      Chirac       & 52 936  & 1 148 208 & 30 148  & 638 547 \\
      Mitterrand   & 8 027   & 218 124   & 5 027   & 134 111 \\        
   \hline
   \end{tabular}
 \caption{Statistiques sur les corpus de développement (D) et de test (T).}
 \label{tab:stats}
 \end{center}
\end{table}

On pourrait être tenté de traiter chacune des trois tâches en appliquant les méthodes employées habituellement en classification. {\itshape A priori}, un problème de classification à deux classes (ici Chirac et Mitterrand\footnote{Pour des facilités d'écriture, nous prenons dorénavant la liberté de désigner les deux derniers présidents de la République, par leur nom de famille, sans les faire précéder d'un titre, ou d'un prénom, et pour plus de concision, il nous arrivera de nous contenter de remplacer « Mitterrand » et « Chirac » par les étiquettes {\itshape M} et {\itshape C}.}) paraît simple. Or, de nombreuses raisons font que la question est complexe. Au terme d'une étude portant sur 68 interventions télévisées composées de 305 124 mots, 
\cite{labbe:l990} distingue quatre périodes dans les discours de Mitterrand. L'une d'elles dénommée « Le président et le premier ministre » (octobre 1986 - mars 1988) n'est probablement pas la plus facile à traiter sous l'angle particulier proposé par le défi DEFT'05. Dans d'autres conditions, c'eût été loin d'être évident. Ici, on peut s'attendre à des difficultés accrues pour différencier deux orateurs qui se sont exprimés dans maints débats sur les mêmes sujets. Facteur aggravant : on ne dispose que d'un petit corpus déséquilibré. Pour la tâche T1, 109 279 mots pleins pour un président et 582 595 pour le second répartis dans 587 discours (dont la date n'est pas fournie). 

Pour donner une idée de la difficulté de ce défi, notons qu'une classification supervisée binaire avec un perceptron optimal à recuit simulé \cite{torres-moreno:2002} appliqué sur la catégorie grammaticale de mots (l'utilisation de tous les mots générant une matrice trop volumineuse) donne un taux d'extraction des segments Mitterrand décevant avec un {\itshape Fscore} $\approx$ 0,43 ; la méthode classique {\itshape K-means} sur les mêmes données, conduit à un {\itshape Fscore} $\approx$  0,4. En comparaison, avec des classifieurs à large marge réputés performants tels que {\itshape AdaBoost} \cite{freund:1997} avec {\itshape BoosTexter} \cite{schapire:2000} et {\itshape Support Vector Machines} avec SVM-Torch \cite{collobert:2001}, on plafonne à un {\itshape Fscore} $\approx$ 0,5. Enfin, avec une méthode de type {\itshape base-line} vraiment simpliste, où on classerait tout segment comme appartenant à la catégorie {\itshape M}, on obtiendrait un {\itshape Fscore} $\approx$ 0,23 sur l'ensemble de développement et de 0,25 pour le test.
Comme ces résultats se sont avérés décevants, nous avons décidé d'explorer des voies totalement différentes. Nous présentons en section \ref{sec:modeles} une approche reposant sur des modèles bayésiens, une chaîne de Markov, des adaptations statiques et dynamiques et un réseau sémantique de noms propres. Des approches probabilistes avec ou sans filtrage et lemmatisation sont utilisées. En section \ref{sec:cohérence}, nous développons une approche de la cohérence interne des discours qui permet encore d'augmenter le {\itshape Fscore}. La section \ref{sec:expériences} est consacrée aux expériences et à leurs résultats. Nous comparons et tentons de fusionner ces différentes approches avant de conclure et d'envisager quelques perspectives. L'annexe présente une analyse détaillée d'un discours de la classe {\itshape C}, où l'utilisation de sa cohérence interne permet de mieux le classer.

\section{Modélisation}
\label{sec:modeles}

La chaîne de traitement que nous allons décrire dans les sous-sections suivantes est constituée de quatre composants : un ensemble de modèles bayésiens (cf. \ref{sec:bayes}), un automate de Markov (cf. \ref{subsec:prise}), un modèle d'adaptation (cf. \ref{subsec:adaptation}) et un réseau sémantique (cf. \ref{subsec:reseau}). Le seul composant totalement dédié à la tâche est l'automate.
Le réseau sémantique dépend du domaine. Sur un Pentium portable cadencé à 1,7 GHz et doté d'une RAM de 384 Mo, l'intégralité de la chaîne d'adaptation s'exécute en 20' qui se décomposent en 5' pour l'apprentissage, et 15' pour le test, soit une minute par itération du couple adaptation-étiquetage. 
Le calcul de la cohérence interne demande un temps supplémentaire de 7' qui reste très raisonnable.

\subsection{Modèles bayésiens}
\label{sec:bayes}

Guidée par une certaine intuition que nous aurions pu avoir des caractéristiques de la langue et du style de chacun des deux orateurs, une analyse des données d'apprentissage aurait pu nous pousser à retenir certaines de leurs caractéristiques plutôt que d'autres. En premier lieu, il aurait été naturel de tabler sur une caractérisation s'appuyant sur les différences de vocabulaire. Des études anciennes comme celles de \cite{cotteret:1969} sur le vocabulaire du Général de Gaulle, ou d'autres plus récentes \cite{labbe:l990} partent du même présupposé. Pour plusieurs raisons, cette approche semble prometteuse mais comme on en rencontre tôt ou tard les limites, on est amené naturellement à ne pas s'en contenter. En effet, la couverture des thématiques abordées par les différents présidents est très large. Il est inévitable que les trajets politiques de deux présidents consécutifs se soient à maintes reprises recoupés. En conséquence, on observe de nombreux points communs dans leurs interventions. On suppose que ces recouvrements viennent s'ajouter les reproductions conscientes ou inconscientes (citations ou effets de mimétisme).

\subsubsection{Modélisation avec lemmatisation}

Au travers d'une modélisation classique \cite{manning:2000}, nous avons testé quelques points d'appuis comme la longueur des phrases (LL), le pourcentage de conjonctions de subordination (Pcos), d'adverbes (Padv) ou d'adjectifs (Padj) et la longueur moyenne des mots pleins (Plm). Cinq de ces variables (Pcos, Padv, Padj, LL, et Plm) ont été modélisées par des gaussiennes $p_i$ dont les paramètres ont été estimés sur le seul corpus d'apprentissage. En ce qui concerne le vocabulaire lui-même, qu'il s'agisse de lemmes ou de mots, nous avons entraîné sur ce même corpus des modèles
{\itshape n}-grammes et {\itshape n}-lemmes (P\#M et P\#L), avec $n < 3$. La probabilité de l'étiquette $t$ (Chirac ou Mitterrand) résulte de la combinaison suivante :

\begin{equation}
\label{eq:lambda}
 P(t)= \sum_{i=1}^r \lambda_i \times p_i(t)
\end{equation}

\noindent avec $\sum_{i=1}^r \lambda_i = 1$. Les valeurs des coefficients $\lambda_i$ que nous avons attribuées de façon empirique à chacune de ces 9 variables $i$ figurent dans le tableau \ref{tab:exemple}. L'estimation de ces valeurs a bien entendu, été réalisée sur le corpus d'apprentissage. Comme le montre le tableau \ref{tab:exemple}, le poids accordé aux lemmes est deux fois plus important que celui accordé aux mots.

\begin{table}[ht]
 \begin{center}
   \tabcolsep = 1.3\tabcolsep
   \begin{tabular}{llllcccccc}
   \hline\hline
        $i$    & P1L & P1M & Padj & LL & P2L & P2M & Pcos & Plm & Padv \\
   \hline
   $\lambda_i$ & 0,30  & 0,15 & 0,15 & 0,15 & 0,30 & 0,15 & 0,05 & 0,02 & 0,01 \\
   \hline
   \end{tabular}
 \caption{Caractères employés pour la modélisation bayésienne et coefficients associés.}
 \label{tab:exemple}
 \end{center}
\end{table}

Lorsqu'on utilise des chaînes de Markov en traitement automatique de la langue naturelle (TALN), on est toujours confronté au problème de la couverture des modèles. Le taux de couverture décroît quand augmente l'ordre du modèle. Le problème est bien connu et des solutions de type lissage ou {\itshape Back-off} \cite{manning:2000} ; \cite{katz:1987} sont une réponse classique au fait que le corpus d'apprentissage ne suffit pas à garantir une estimation fiable des probabilités. Le problème devenant critique lorsqu'il y a un déséquilibre flagrant entre les deux classes, il nous a semblé inutile, voire contre-productif de calculer des tri-grammes.

En nous inspirant des travaux menés en lexicologie sur les discours de Mitterrand, nous avons essayé de prendre en compte certains des traits qualifiés de dominants chez Mitterrand par \cite{illouz:2000} : adverbe négatif, pronom personnel à la première personne du singulier, point d'interrogation, ou des expressions comme « c'est », « il y a », « on peut », « il faut » (dans les quatre cas, à l'indicatif présent). Ceux-ci ont été traités de la même façon que les autres caractères de la modélisation bayésienne. Après vérification de la validité statistique de ces traits sur le corpus DEFT'05, nous les avons intégrés dans la modélisation mais dans un second temps, nous les avons retirés car même s'ils entraînaient une légère amélioration sur les données de développement, rien ne garantissait qu'il ne s'agissait pas, là, de tics de langage liés à une période potentiellement différente de celle du corpus de test. Par ailleurs, en cas de portage de l'application à un autre domaine ou une autre langue, nous ne voulions pas être dépendants d'études lourdes. En tous les cas, nous avons préféré faire confiance aux modèles de Markov pour capturer automatiquement une grande partie de ces tournures.

\subsubsection{Modélisation sans lemmatisation}

Parallèlement et à l'inverse de nos préoccupations de la sous-section précédente, nous avons souhaité faire fonctionner nos modèles sur le texte à l'état brut, sans enrichissement ou annotation. Pour aller dans ce sens, nous nous sommes demandé à quel point la recherche automatisée des caractéristiques propres à un auteur pourrait être facilitée ou perturbée par le
fait de ne pas filtrer ni éliminer quoi que ce soit des discours. Ainsi, nous avons fait l'hypothèse que
l'utilisation répétée, voire exagérée de certains termes ne servant qu'à assurer le bâti de la phrase, pouvait
prétendre au statut d'indicateur fiable. Pour ce deuxième modèle\footnote{Qui sera appelé par la suite Modèle II et par opposition celui avec lemmatisation sera appelé Modèle I.}, nous sommes partis du principe que les
techniques de {\itshape n}-gra\-mmes appliquées à des tâches de classification, pourraient se passer d'une phase
préalable de lemmatisation ou de {\itshape stemming}, du rejet des mots-outils et de la ponctuation. Les systèmes {\itshape
n}-grammes, \cite{jalam:2002} ; \cite{sahami:1999} ont montré que leurs performances ne s'améliorent pas après
{\itshape stemming} ou élimination des mots-outils. Dans cet esprit, nous avons laissé les textes dans leur état
originel. Aucun prétraitement n'a été effectué, même si cette démarche a ses limites : par exemple, «~Gasperi~»
et « Gaspéri » comptent pour des mots différents, qu'il y ait ou non erreur d'accent ; « premier » et « première
» sont aussi comptabilisés séparément en absence de lemmatisation. Malgré cela, nous avons voulu donner au
modèle un maximum de chances de capturer des particularités de style (manies de ponctuer le texte par l'emploi
de telle ou telle personne, de subjonctifs, gérondifs,\ldots) qui sont gommées après application de
certains prétraitements comme la lemmatisation. Une classification naïve et un calcul d'entropie ont déjà été rapportés
lors de l'atelier DEFT'05 avec un automate légèrement différent \cite{el-beze:2005}. Seule variante, l'ajout d'une contrainte : tout mot de longueur $\leq$ 5 n'est pas pris en compte afin d'alléger les calculs. Ceci correspond à un « filtrage » relativement indépendant de la langue.

\subsection{Automate de Markov}
\label{subsec:prise}

Comme cela était dit en introduction, un discours de Chirac peut avoir fait l'objet de l'insertion d'au plus une séquence de phrases. La séquence {\itshape M}, si elle existe, est d'une longueur supérieure ou égale à deux. Pour prendre en compte cette contrainte particulière, nous avions, initialement, pensé écrire des règles, même si une telle façon de faire
s'accorde généralement peu avec les méthodes probabilistes. Dans le cas présent, que faut-il faire si une phrase
détachée de la séquence {\itshape M} a été étiquetée {\itshape M}, avec une probabilité plus ou moins élevée
(certainement au dessus de 0,5, sinon elle aurait reçu l'étiquette {\itshape C}) ? Renverser la décision, ou la
maintenir ? Si l'on opte pour la seconde solution, il serait logique d'extraire également toutes les phrases qui
la séparent de la séquence {\itshape M}, bien qu'elles aient été étiquetées {\itshape C}. Mais, dans ce cas, un gain aléatoire en rappel risque de se faire au prix d'une chute de précision.

\begin{figure}[ht]
\begin{center}
\resizebox{0.75\columnwidth}{!}{%
 \includegraphics{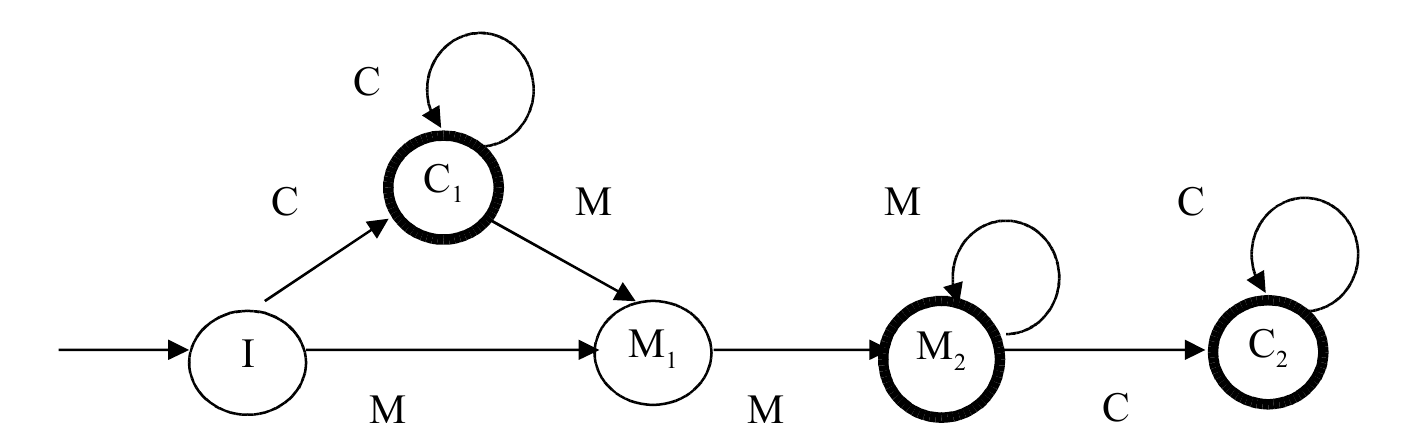}
 }
 \caption{Machine de Markov exprimant les contraintes générales des trois tâches.}
 \label{fig:markov}
\end{center}
\end{figure}

Pour pouvoir trouver, parmi les chemins allant du début à la fin du discours, celui qui optimise la production
globale du discours, nous avons exploité un automate probabiliste à cinq états (dont un initial {\itshape I} et
trois terminaux, $C_1$, $C_2$, et $M_2$. Comme on peut le voir sur la figure
\ref{fig:markov}, vers les états dénommés $C_1$ et $C_2$ (respectivement $M_1$ et
$M_2$) n'aboutissent que des transitions étiquetées {\itshape C} (respectivement {\itshape M}). À une
transition étiquetée {\itshape C} (respectivement {\itshape M}), est associée la probabilité d'émission
combinant pour {\itshape C} (respectivement {\itshape M}) les modèles probabilistes définis en section
\ref{sec:bayes}.

Avant de décrire les étapes ultérieures du processus de catégorisation segmentation, notons que c'est ce
composant qui a permis de faire un saut conséquent (plus de 25\% en absolu) au niveau des performances et a
ouvert ainsi la voie à la mise en place de procédures d'adaptation décrites en section suivante. S'il s'avère
qu'étiqueter un bloc de plusieurs segments est plus fiable qu'étiqueter individuellement chaque phrase, il est
naturel que cela ait un impact positif  sur les performances.

Remarquons par ailleurs que la question aurait pu être gérée autrement, par exemple en utilisant, pour chaque
discours, la partie triangulaire supérieure d'une matrice carrée  $\Psi[d,d]$ ($d$ étant le nombre de phrases
contenues dans le discours en question, voir les figures \ref{fig:decoupage} et \ref{fig:matrice}). Dans chaque
case $\Psi[i,j]$, on calcule la probabilité que la séquence soit étiquetée {\itshape M} entre $i$ et $j$, et
{\itshape C} du début jusqu'au $i-1$ et de $j+1$ à $d$. Déterminer les bornes optimales de la séquence
Mitterrand revient alors à rechercher un maximum sur toutes les valeurs $\Psi[i,j]$ telles que $i > j$. Si cette valeur optimale est inférieure à celle qu'on aurait obtenue en produisant toute la chaîne avec le modèle associé
à Chirac, on se doit de supprimer la séquence {\itshape M}. Sauf si on factorise les calculs pour remplir les différentes cases,
la complexité de cette seconde méthode est supérieure à celle de l'algorithme de Viterbi \cite{manning:2000}. Il nous a paru néanmoins intéressant d'en faire état dès à présent, car elle offre la possibilité de combiner aisément des
contraintes globales plus élaborées que celles que nous prenons en compte dans l'adaptation. Elle peut aussi
permettre de mixer des modèles issus de l'apprentissage et d'autres optimisant des variables dédiées à la
modélisation de la cohésion interne des séquences qui se trouvent dans le discours traité, et n'ont fait l'objet
d'aucun apprentissage préalable, comme nous le montrerons en section \ref{sec:cohérence}.

\subsection{Adaptation statique et dynamique}
\label{subsec:adaptation}

La contrainte de ne pouvoir enrichir le corpus d'apprentissage, sous peine de disqualification\footnote{« Les
équipes utilisant dans leur méthode des corpus de J. Chirac et de F. Mitterrand autres que ceux fournis par les
organisateurs seront disqualifiées. Par exemple, la méthode consistant à acquérir un corpus de F. Mitterrand
et/ou de J. Chirac par des ressources extérieures pour identifier les phrases de F. Mitterrand présentes dans le
corpus fournis par les organisateurs sera considérée comme non valide. » Source :
http://www.lri.fr/ia/fdt/DEFT05/resultats.html}, nous a poussé à tirer un parti intégral des données mises à
notre disposition. Or, en dehors du corpus d'apprentissage, il ne restait plus qu'une issue : intégrer dans
l'apprentissage (bien entendu, sans les étiquettes de référence) une partie des données de test. C'est sur
ces données que l'adaptation a été pratiquée. Dériver un modèle à partir de l'intégralité des discours de test
correspond à ce que nous appelons ici {\itshape adaptation statique}. L'{\itshape adaptation dynamique}, quant à elle, repose sur un
modèle découlant seulement du discours en train d'être testé. \'Evidemment, il n'est pas interdit de
conjuguer les deux approches.

Dans un premier temps, nous avions envisagé de pratiquer un étiquetage des données de test, l'objectif étant à
l'itération $i+1$ de n'adjoindre au corpus d'apprentissage\footnote{$X$ pouvant prendre ici les valeurs
{\itshape M} ou {\itshape C}.} de $X$ que les phrases $s$ ayant reçu au pas $i$ une probabilité $P_i(X|s)$
supérieure à un certain seuil $T_i$. Un apprentissage de type maximum de vraisemblance effectué sur les données
ainsi collectées peut autant rapprocher qu'éloigner du point optimal. Pour pallier cette difficulté, nous avons
opté pour un apprentissage d'{\itshape Expectation-Maximisation} (EM), consistant à ne compter pour chaque
couple \{ élément $= e, X \}$ observé dans les données d'adaptation que la fraction d'unité égale à la
probabilité de l'orateur $X$ sachant la phrase qui contient $e$. La prise de décision repose sur une formule
analogue à celle de la formule (\ref{eq:lambda}). La variable en position 0 est la probabilité de l'étiquette
sachant la phrase qui lui a été attribuée à l'itération $i$. Nous avons fait décroître le poids $\lambda_0$ qui
lui est associé, de façon progressive, d'une itération à l'autre par pas de 0,1. Les quatre modèles employés
sont, pour les deux premiers, lemmes et mots issus de l'adaptation locale (dynamique), pour les deux derniers,
lemmes et mots issus de l'adaptation globale (statique). La pondération entre les différentes probabilités est
restée la même durant toutes les itérations : Dynamique \{ lemmes = 0,4 ; mots = 0,1  \} / Statique \{ lemmes = 0,4 ; mots = 0,1 \}. Les procédures d'adaptation statique et dynamique mises en \oe uvre durant cette étape ont permis de gagner entre 3 et 4 points de {\itshape Fscore}.

\subsection{Réseau de Noms Propres}
\label{subsec:reseau}

À partir de la tâche T2, l'ensemble des noms propres était dévoilé aux participants. Établir un lien entre
différents éléments apparaissant dans des phrases même éloignées d'un discours donné, nous a paru être un bon
moyen pour mettre en évidence une sorte de réseau sémantique permettant aux segments de s'auto-regrouper autour
d'un lieu, de personnes et de façon implicite d'une époque. Dans le cas de données bien séparables, plusieurs
ensembles de noms ancrés dans une Histoire et une Géographie commune devraient former des composantes connexes
(idéalement deux) sur lesquelles il suffirait ensuite de mettre l'étiquette {\itshape M} ou {\itshape C}. Bien
que cela ne soit pas tout à fait la démarche que nous avons adoptée, ces remarques aident à en comprendre l'esprit.

2 171 termes ont été regroupés dans 314 « concepts »\footnote{Les termes ont été regroupés de façon
manuelle pour former les concepts du réseau.} qui pour épouser la richesse des discours traités dépassent largement un cadre restreint aux seules
considérations géopolitiques (le Sport et la Culture sont souvent abordés lors de cérémonies de remises de
médailles). Un terme peut se retrouver dans plusieurs classes, comme par exemple «~ Miguel Angel
Asturias~», qui a été placé aussi bien dans la classe des Guatémaltèques que dans celle des écrivains
étrangers. Afin de mixer les relations entretenues entre les noms de pays, leurs habitants, les capitales, le
pouvoir exécutif, nous avons complété un réseau fourni par le Centre de Recherche de
Xerox\footnote{http://www.xrce.xerox.com}, en y rajoutant quelques relations issues des Bases de Connaissance
que l'équipe TALN du LIA utilise pour faire fonctionner son système de Questions / Réponses \cite{bellot:2003}. En table \ref{tab:extrait}, figure un petit extrait de ce réseau non structuré.
\begin{table}[ht]
 \begin{center}
   \tabcolsep = 1.6\tabcolsep
   \begin{tabular}{cl}
   \hline\hline
    CONCEPT & TERMES \\
   \hline
   Argentin      & Argentine Alfonsin Carlos\_Menem Bioy\_Casares Buenos\_Aires \\
                 & Alfredo\_Arias Jorge\_Remes \\
   \hline
   Guatemalteque & Guatemala\_Ciudad Guatemala Guatémaltèque Guatémaltèques \\
                 & Guatémaltais Guatémaltaise Guatémaltaises Ciudad\_Vieja Permedo \\
                 & \textbf{Miguel\_Angel\_Asturias} Alvaro Arzu Irigoyen Rigoberta\_Menchu \\
                 & Rigoberta \\
   \hline
   Mexicain      & Mexique Mexico Zedillo Zédillo Benito\_Juarez Carlos\_Fuentes \\
                 & Octavio\_Paz FOX Fox Cancun Monterrey Mexicain Mexicaine \\
                 & Mexicaines Mexicains \\
   \hline
   Ecrivains\_e   & Gao Saramago Virgilio\_Ferreira Fernando\_Pessoa Ionesco Cioran \\
                  & Fukuzawa\_Yukichi Nadia\_Tueni Amin\_Maalouf Tahar\_Ben\_Jelloun \\
                  & Senghor Rachid\_Boujedra Bioy\_Casares Boubou\_Hama \\
                  & Hector\_Bianciotti \textbf{Miguel\_Angel\_Asturias} Dostoïevsky \\
   \hline
   \end{tabular}
 \caption{Extrait du réseau de noms propres.}
 \label{tab:extrait}
 \end{center}
\end{table}
On enrichit les segments en leur ajoutant les concepts auxquels appartiennent les termes qui les composent. Deux segments qui ont en commun plus d'un certain nombre d'éléments (termes ou concepts) sont considérés 2 à 2 comme liés thématiquement et mis dans une même classe de segments. Les classes sont élargies par une itération de ce processus. La probabilité de chaque segment est combinée avec la probabilité de la classe à laquelle il appartient.
Après quatre itérations, sur 57 301 phrases valides que comptait le corpus de développement (test : 27 120), 6
285 ont été regroupées en 942 groupes (test : 456 groupes de 3 127). Un peu plus de 11\% des segments se
retrouvent donc dans des groupes, dont le cardinal moyen est d'environ 7 phrases. Le plus grand groupe contient
50 segments (test : 66). Seuls 16 groupes (test : 12) regroupent, de façon confuse, des étiquettes {\itshape M}
et {\itshape C}. C'est le cas du discours 38, où la phrase 30 étiquetée {\itshape M} possède en commun «
Casablanca MAGHREB » (en fait, il s'agissait du sommet de Casablanca) avec la phrase 173 étiquetée {\itshape C},
où Chirac fait état de ses récents voyages au Maroc. L'avantage d'un réseau probabiliste est que cette erreur
n'est pas rédhibitoire. En effet, dans notre soumission, la phrase 30 a été correctement extraite et non la
phrase 173. Cela ne fonctionne pas toujours aussi bien ! Dans le cas du discours 739, la séquence {\itshape C}
et la séquence {\itshape M} ont en commun deux « termes-concepts » (« Espagne-Espagnol » et «
Méditerranée-Méditerranéen »). Il se trouve que la seconde confusion aurait pu être évitée si le TGV
Paris-Lyon-Méditerranée dont parle Mitterrand n'avait pas fait l'objet d'une sur-découpe au moment de la
{\itshape tokenisation}. Mais cela n'aurait pas suffi, car avec l'aide de l'autre terme (Espagne) quatre phrases {\itshape
M} (30, 35, 36 et 37) ont été regroupées par transitivité avec 12 phrases étiquetées {\itshape C} (1, 3, 6-17,
20-25, 27, 47). De fait, aucun segment du discours 739 n'a été extrait. Il est clair que nous sommes encore loin
d'une représentation élaborée des relations entretenues entre des concepts et leur expression au travers de
textes. Néanmoins, le réseau que nous avons élaboré à peu de frais est un premier pas dans cette direction.

\section{Cohésion thématique des discours}
\label{sec:cohérence}

En section \ref{subsec:prise}, nous avions avancé l'hypothèse qu'étiqueter un bloc est plus fiable qu'étiqueter chaque phrase de façon indépendante l'une de l'autre. Cela se discute en fait si on se borne à rechercher la suite de segments qui optimise la cohésion thématique\footnote{La cohésion thématique étant un des éléments permettant d'apprécier la cohérence interne d'un discours, nous emploierons de préférence l'expression « cohérence interne » dans la suite de l'article.} de chacun des deux blocs, il est indispensable de conjuguer
cette approche thématique avec un étiquetage d'auteur.
Cette étape est motivée par une idée simple découlant des présupposés de DEFT'05 :
« \emph{Les passages de F. Mitterrand introduits traitent d'une thématique différente. Par exemple, dans les
allocutions de J. Chirac évoquant la politique internationale, les phrases de F. Mitterrand introduites sont
issues de discours traitant de politique nationale. Ainsi, la rupture thématique peut être une des manières de
détecter les phrases issues du corpus de F. Mitterrand.} »\footnote{Source : http://www.lri.fr/ia/fdt/DEFT05}

Dans cette optique, on peut vouloir trouver un découpage de chaque discours soit en un bloc ({\itshape
C\ldots C}), soit en deux blocs ({\itshape C \ldots C-M \ldots M}) ou ({\itshape M\ldots M-C\ldots C}) soit en trois
blocs ({\itshape C\ldots C-M\ldots M-C\ldots C}) tels que le bloc des segments étiquetés {\itshape M} ou
les blocs (1 ou 2) étiquetés {\itshape C} présentent tous les deux une cohérence thématique interne optimale.
Pour cela, nous proposons de formaliser le problème comme suit : la probabilité de production d'une phrase est
évaluée au moyen d'un modèle appris sur toutes les phrases du bloc auquel elle appartient sauf elle. En
maximisant le produit des probabilités d'émission de toutes les phrases du discours, on a toutes les chances de
bien identifier des ruptures thématiques. Mais rien ne garantit qu'elles correspondent à des changements d'orateurs.
En effet, supposons qu'il n'y ait pas dans un discours donné, d'insertion de phrases de Mitterrand, et que dans
les discours de Chirac se trouve une longue digression de 20 phrases. Notre approche risque de reconnaître à
tort ces 20 phrases comme attribuables à la classe {\itshape M}. Pour éviter ce travers, nous proposons une
optimisation mettant en \oe uvre conjointement les modèles de cohérence interne et ceux issus de l'apprentissage.
En annexe, nous donnons en exemple le discours 520 pour lequel ce phénomène se produit. Nous voyons comment la
cohérence interne réussit à renverser presque totalement la situation : ainsi, un gros bloc étiqueté {\itshape
M} par l'adaptation seule, au sein d'un discours dont la classe est {\itshape C} a été étiqueté correctement par
la cohérence interne, à l'exception de deux phrases dont les probabilités penchaient trop fortement vers la
classe {\itshape M}.

\begin{figure}[ht]
\begin{center}
 \includegraphics[width=7cm]{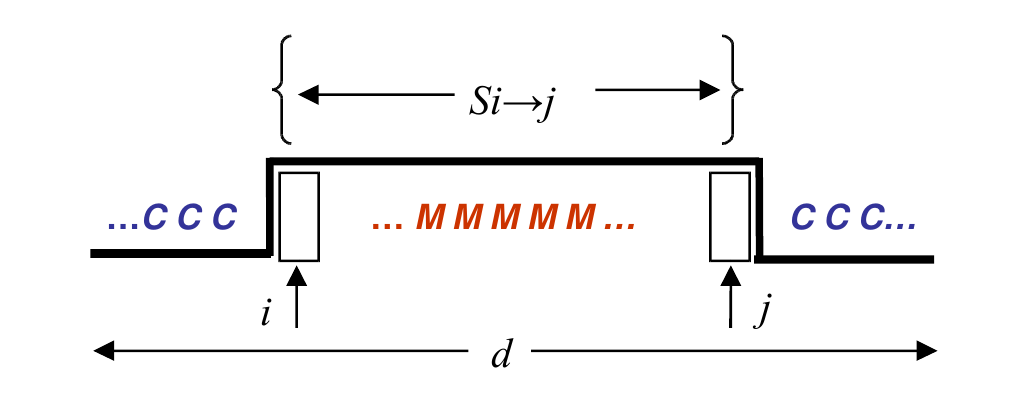}
 \caption{Schéma de découpage des discours.}
 \label{fig:decoupage}
\end{center}
\end{figure}

Le modèle de cohérence interne cherche donc à maximiser la probabilité d'appartenance des phrases proches aux
frontières de segments. Il peut utiliser \emph{a)} le réseau de noms propres et \emph{b)} la probabilité issue
de l'apprentissage par Markov. Pour un discours $S^{d}_1$ donné, de longueur {\itshape d}, nous cherchons un
découpage optimal $\widetilde{D}$ (cf. figure \ref{fig:decoupage}) et un étiquetage $\widetilde{E}$ tels que :

\begin{equation}
(\widetilde{D},\widetilde{E}) = \textrm{Arg}\max_{D,E} \left\{P_{I}(D,E|S^{d}_{1}) \times P'(D,E|S^{d}_{1})\right\}
\end{equation}

\noindent où $P'(D,E|S^{d}_{1})$ est la probabilité issue de l'apprentissage et $P_{I}(D,E|S^{d}_{1})$ la probabilité de
cohérence interne (à l'intérieur d'un discours). La conjugaison des modèles d'apprentissage et de cohérence interne est
réalisée par le produit entre $P'$ et $P_{I}$, qu'il semble légitime de considérer indépendants l'un de l'autre. Comme le découpage ne peut être déduit de l'apprentissage, nous
faisons l'hypothèse que $P'(D,E|S^{d}_{1}) \cong P'(E|S^{d}_{1})$. Donc :

\begin{equation}
\label{eq:DE}
(\widetilde{D},\widetilde{E}) = \textrm{Arg}\max_{D,E} \left\{P_{I}(D,E|S^{d}_{1}) \times P'(E|S^{d}_{1})\right\}
\end{equation}

\noindent Or, d'après le théorème de Bayes :

\begin{equation}
P_{I}(D,E|S^{d}_{1}) = \frac{P(S^{d}_{1}|D,E)P(D|E)}{P(S^{d}_{1})}  \mathrm{\ \  et \ \ }  P'(E|S^{d}_{1}) =
\frac{P'(S^{d}_{1}|E)P'(E)}{P'(S^{d}_{1})}
\end{equation}

\noindent De ce fait, l'équation (\ref{eq:DE}) devient :

\begin{equation}
(\widetilde{D},\widetilde{E}) \cong \textrm{Arg}\max_{D,E} \left\{\frac{P(S^{d}_{1}|D,E)P(D|E)}{P(S^{d}_{1})} \times
\frac{P'(S^{d}_{1}|E)P'(E)}{P'(S^{d}_{1})}\right\}
\end{equation}
\noindent Nous savons que $P(D|E)$ prend toujours des valeurs \{0, 1\} car le découpage est toujours déterminé par les
étiquettes (mais pas vice-versa). La probabilité $P'(E)$ ne peut pas être déduite de l'apprentissage (le choix
de $D$ peut être considéré comme aléatoire) et $P(S)$ et $P'(S)$ ne dépendent pas de $D$ ou de $E$. Alors :
\begin{equation}
(\widetilde{D},\widetilde{E}) \cong \textrm{Arg}\max_{D,E} \left\{ P_{I}(S^{d}_{1}|D,E) \times P'(S^{d}_{1}|E)\right\}
\end{equation}

\noindent Nous avons choisi de représenter un couple $(D,E)$ par un couple de deux indices $i$ et $j$ dont la signification 
est donnée par la figure \ref{fig:decoupage}. Ces deux indices correspondent aux bornes du bloc des segments étiquetés $M$ et à la ligne et la colonne de la matrice $\Psi$ évoquée en section \ref{subsec:prise}. 
Pour un discours donné, on aura donc :
\begin{equation}
\Psi[i,j] = P(S_{1 \ldots i-1,j+1 \ldots d}|C) \times P(S_{i \ldots j}|M) \times P'(S_{1 \ldots i-1,j+1 \ldots
d}|C) \times P'(S_{i \ldots j}|M)
\end{equation}
\noindent En faisant l'hypothèse\footnote{Cette hypothèse va quelque peu à l'encontre de l'objectif recherché, à savoir
considérer les segments d'un même bloc comme un tout, mais nous ne savons pas comment faire autrement.} que les
segments sont indépendants, nous introduisons le produit sur toutes les phrases du discours en distinguant
celles qui sont à l'intérieur du bloc $S_{i \rightarrow j} (k=i \ldots j)$ de celles qui sont à l'extérieur
$(k=1 \ldots i-1, j+1 \ldots d)$ :
\begin{equation}
\Psi[i,j] = \prod\limits_{ k=1 \ldots i-1,j+1 \ldots d }[P(S_{k}|\chi) \times P'(S_{k}|C)] \times
\prod\limits_{k=i \ldots j}[P(S_{k}|\mu) \times P'(S_{k}|M)]
\end{equation}
\noindent où $d$ est la longueur du discours et $\chi = C \setminus S_{k}$ et $\mu = M \setminus S_{k}$. Ceci revient à exclure le segment $S_k$ des données qui servent à estimer les paramètres utilisés pour calculer la probabilité de production de ce même segment $S_k$.  Notons que, si les probabilités
$P(S_{k}|\chi)=1$ et $P(S_{k}|\mu)=1$, alors la valeur de $\Psi[i,j]$ est réduite au cas de Markov (adaptation simple).
Nous avons exploité la matrice $\Psi[i,j]$ (cf. figure \ref{fig:matrice}) en nous limitant à sa partie triangulaire
supérieure. Le fait d'exclure la diagonale principale dans les calculs illustre l'exploitation de la contrainte respectée par les fournisseurs du corpus DEFT'05. S'il y a des segments de la classe {\itshape M} insérés, il y a en au moins deux. Le cas des discours étiquetés uniquement {\itshape C} n'est pas représenté dans la figure, mais il a été pris en considération, même s'il n'est pas intégré dans la matrice.

\begin{figure}[ht]
 \begin{center}
    \tabcolsep = 2.6\tabcolsep
 \begin{tabular}{|c|c|c|c|c|c|c|}
 \hline
 1 & & ... & & $i$ & ... & $d$ \\
 \hline
 $\vdots$ & $\bullet$ & & & $\vdots$ & & \\
 \hline
 $j$      & $\bullet$ & $\bullet$ & & $P(S_i,S_j)$ & & \\
 \hline
          & $\bullet$ & $\bullet$ & $\bullet$ & $\vdots$ & & \\
 \hline
 $\vdots$ & $\bullet$ & $\bullet$ & $\bullet$ &$\bullet$ & & \\
 \hline
          & $\bullet$ & $\bullet$ & $\bullet$ & $\bullet$ &$\bullet$ &  $\vdots$ \\
 \hline
 $d$ & $\bullet$ & $\bullet$ & $\bullet$ & $\bullet$ & $\bullet$ & $\bullet$ \\
 \hline
 \end{tabular}
 \caption{Matrice $\Psi[i,j]$ pour le calcul de la cohérence interne. 
          Les  $\bullet$ représentent les cases ignorées pour le calcul des probabilités.}
 \label{fig:matrice}
 \end{center}
 \end{figure}

\section{Expériences}
\label{sec:expériences}

Pour la Modélisation I, tous les corpus (apprentissage et test) ont été traités par l'ensemble d'outils
LIA\_TAGG\footnote{Téléchargeable à l'adresse : http://www.lia.univ-avignon.fr}. Ces outils contiennent les modules suivants :

\begin{itemize}
	\item un module de formatage de texte permettant de découper un texte en unités (ou {\itshape tokens}) en accord avec un lexique de référence ;
  \item un module de segmentation insérant des balises de début et fin de phrase dans un flot de texte, en accord avec un certain nombre d'heuristiques ;
  \item un étiqueteur morphosyntaxique, basé sur l'étiqueteur ECSTA \cite{spriet:1998} ;
  \item un module de traitement des mots inconnus permettant d'attribuer une étiquette morphosyntaxique à une forme inconnue du lexique de l'étiqueteur en fonction du suffixe du mot et de son contexte d'occurrence. Ce module est basé sur le système DEVIN présenté dans \cite{spriet:1996}.
  \item un lemmatiseur associant à chaque couple mot/étiquette morphosyntaxique un lemme en fonction d'un lexique de référence.
\end{itemize}

Dans la phase de
développement, le corpus d'apprentissage a été découpé en cinq sous-corpus de telle sorte que pour chacune des
cinq partitions, un discours appartient dans son intégralité soit au test soit à l'apprentissage. À tour de
rôle, chacun de ces sous-corpus est considéré comme corpus de test tandis que les quatre autres font office de
corpus d'apprentissage. Cette répartition a été préférée à un tirage aléatoire des phrases tolérant le
morcellement des discours. En effet, un tel tirage au sort présente deux inconvénients majeurs. Le premier
provient du fait qu'un tirage aléatoire peut placer dans le corpus de test des segments très proches de segments
voisins qui eux ont été placés dans le corpus d'apprentissage. Le second inconvénient (le plus gênant des deux),
tient au fait qu'une telle découpe ne permet pas de respecter le schéma d'insertion défini dans les spécificités de
DEFT'05.

\subsection{Résultats de l'adaptation}

\begin{figure}[ht]
\begin{center}
 \includegraphics[width=6.5cm]{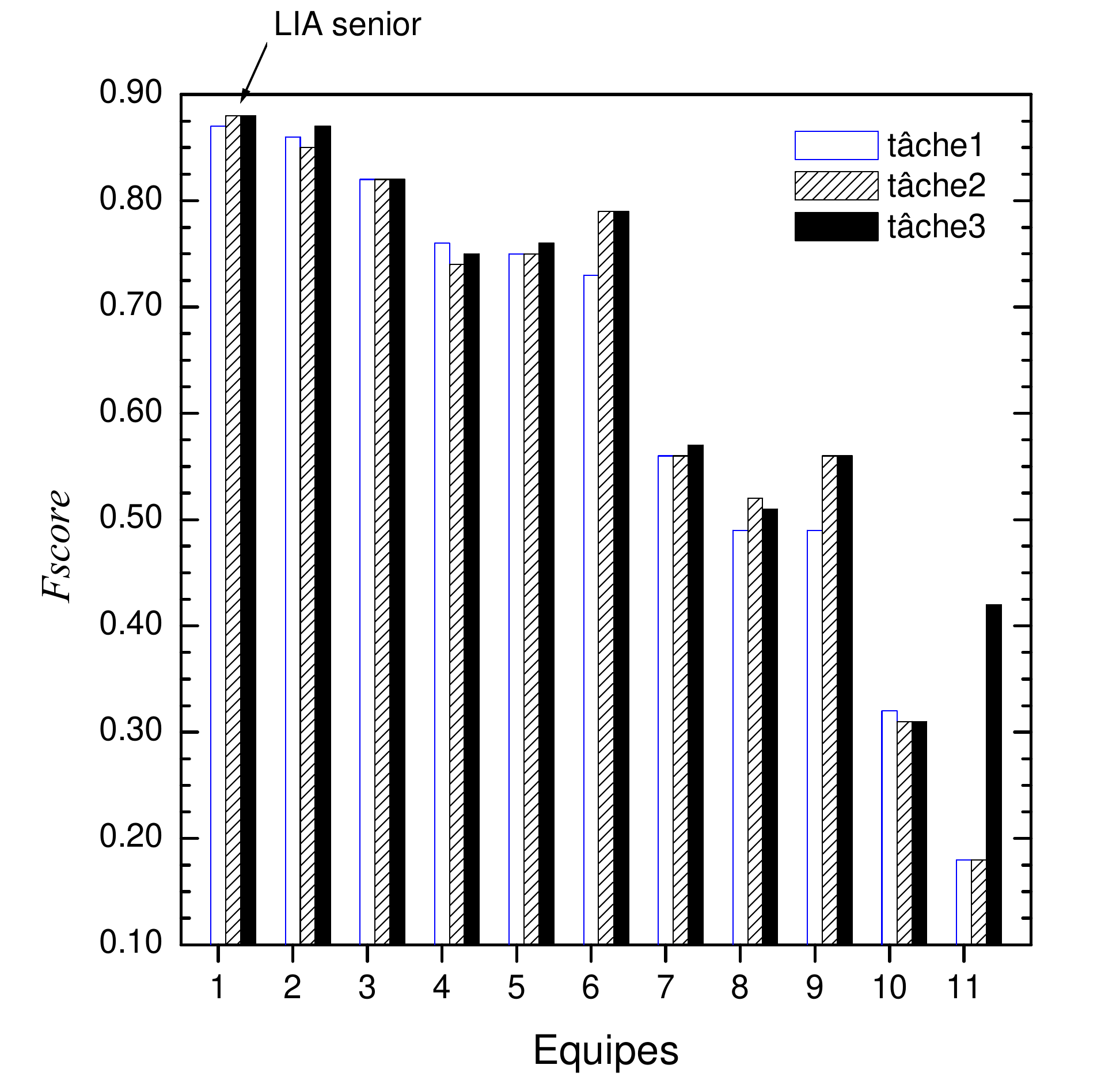}
 \caption{{\itshape Fscore} officiels pour les trois tâches (T1 : pas de noms, pas de dates ; T2 : pas de noms et T3 : avec noms et dates) pour l'ensemble de participants DEFT'05. Les membres des équipes sont cités au tableau \ref{tab:resultats}.}
 \label{fig:officiels}
\end{center}
\end{figure}

Des résultats de nos modèles uniquement avec adaptation ont été publiés dans les actes du colloque TALN 2005.
Nous reproduisons ici les observations majeures qui pouvaient être faites sur ces résultats. Le {\itshape
Fscore} s'améliore de façon notable au cours des cinq premières itérations de l'adaptation. Au-delà, il n'y a
pas à proprement parler de détérioration mais une stagnation qui peut être vue comme la captation par un maximum
local. L'apport des réseaux bâtis autour des noms propres est indéniable \cite{el-beze:2005}. Nous montrons
au tableau \ref{tab:resultats} et sur la figure \ref{fig:officiels} les meilleurs {\itshape Fscore} officiels
soumis pour l'ensemble de participants pour les trois tâches. On peut voir que le système du LIA senior est
positionné, dans les trois cas, en première place. La méthode de \cite{rigouste:2005} en deuxième position,
utilise quelques méthodes probabilistes semblables aux nôtres, mais ils partent de l'hypothèse où la segmentation thématique est faite au niveau des orateurs (pas au niveau du discours), ils ont besoin
de pondérer empiriquement les noms et les dates (tâches T2 et T3), leurs machines de Markov sont plus complexes et il ne font pas  d'adaptation, entre autres. 

Le dévoilement des dates (tâche T3) permet d'améliorer très légèrement les résultats du modèle II, mais entraîne une dégradation sur le modèle I. En ce qui concerne la précision et le rappel au fil des itérations sur l'ensemble de Test(T) ainsi que sur le Développement(D), c'est le gain en précision qui explique l'amélioration due aux Noms Propres. Ce gain
allant de pair avec un rappel quasi identique (légèrement inférieur pour le test), il apparaît que le composant
Noms Propres fonctionne comme un filtre prévenant quelques mauvaises extractions (mais pas toutes).

\begin{table}[ht]
 \begin{center}
   \tabcolsep = 1.6\tabcolsep
   \begin{tabular}{lccc}
   \hline\hline
    Equipes & T1 & T2 & T3 \\
   \hline
    \textbf{1 El-Bèze, Torres, Bechet : LIA senior} & \textbf{0,87} & \textbf{0,88} & \textbf{0,88} \\
    2  Rigouste, Cappé, Yvon : ENST & 0,86 & 0,85 & 0,87 \\
    3  Pierron, Durkal, Freydigier : LORIA/UHP & 0,82 & 0,82 & 0,82 \\
    4  Labadie, Romero, Sitbon : LIA junior & 0,76 & 0,74 & 0,75 \\
    5  Maisonnasse, Tambellini : CLIPS & 0,75 & 0,75 & 0,76 \\
    6  Kerloch, Gallinari : LIP6 & 0,73 & 0,79 & 0,79 \\
    7  Hernandez : LIMSI & 0,56 & 0,56 & 0,57 \\
    8  Plantié, Dray, Montmain, Meimouni, Poncelet : LGI2P & 0,49 & 0,52 & 0,51 \\
    9  Hurault-Plantet, Jardino, Illouz : LIMSI & 0,49 & 0,56 & 0,56 \\
    10 Chauche : LIRMM & 0,32 & 0,31 & 0,31 \\
    11 Henry, Marley, Amblard, Moot : LABRI & 0,18 & 0,18 & 0,42 \\
   \hline\hline
   \end{tabular}
 \caption{Résultats officiels de l'atelier DEFT'05 du {\itshape Fscore} sur les trois tâches de test \{T1, T2, T3\} pour les meilleures soumissions de l'ensemble des 11 équipes.}
 \label{tab:resultats}
 \end{center}
\end{table}

\subsection{Résultats avec la cohérence interne}

\begin{figure}[ht]
\begin{center}
 \includegraphics[width=6cm]{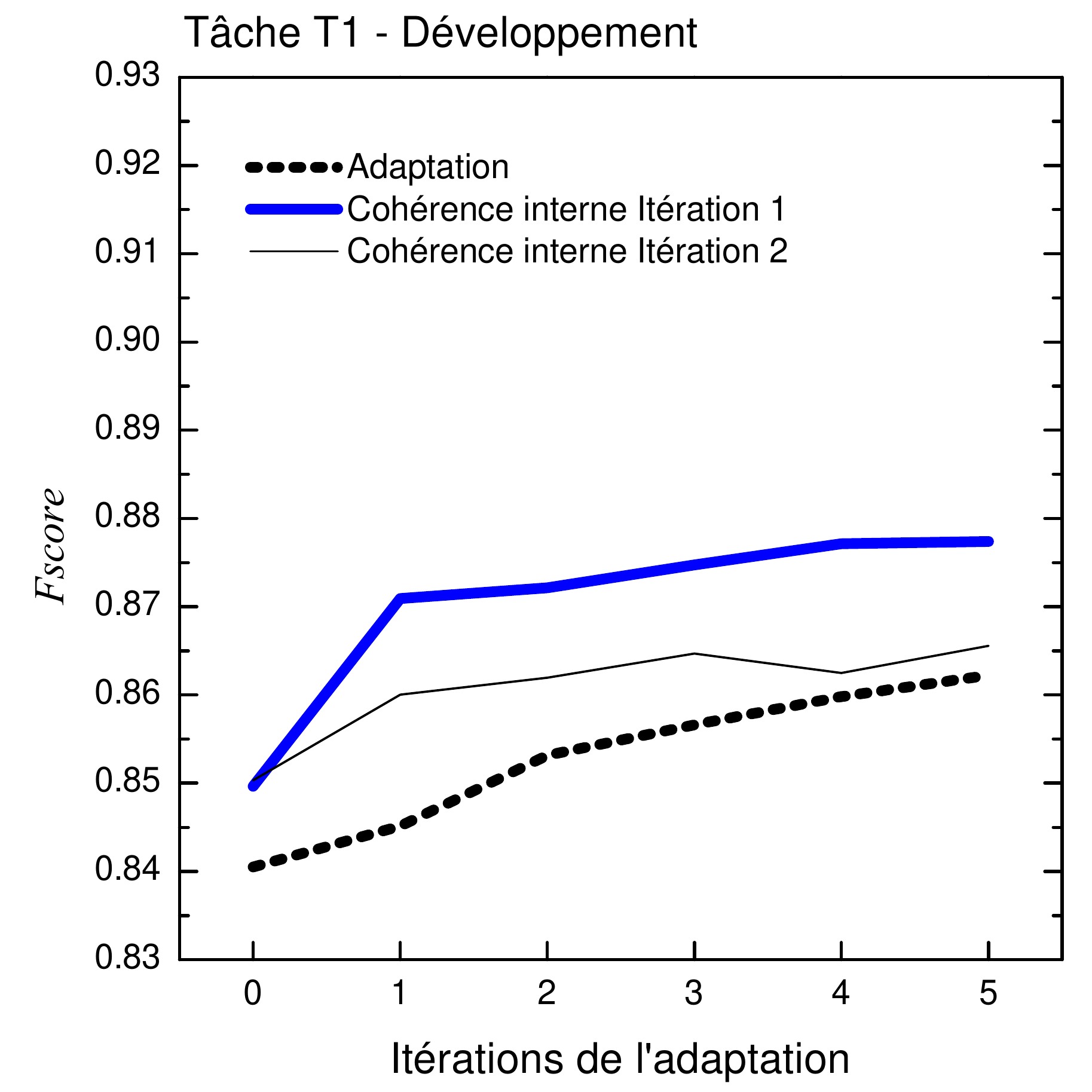}
 \includegraphics[width=6cm]{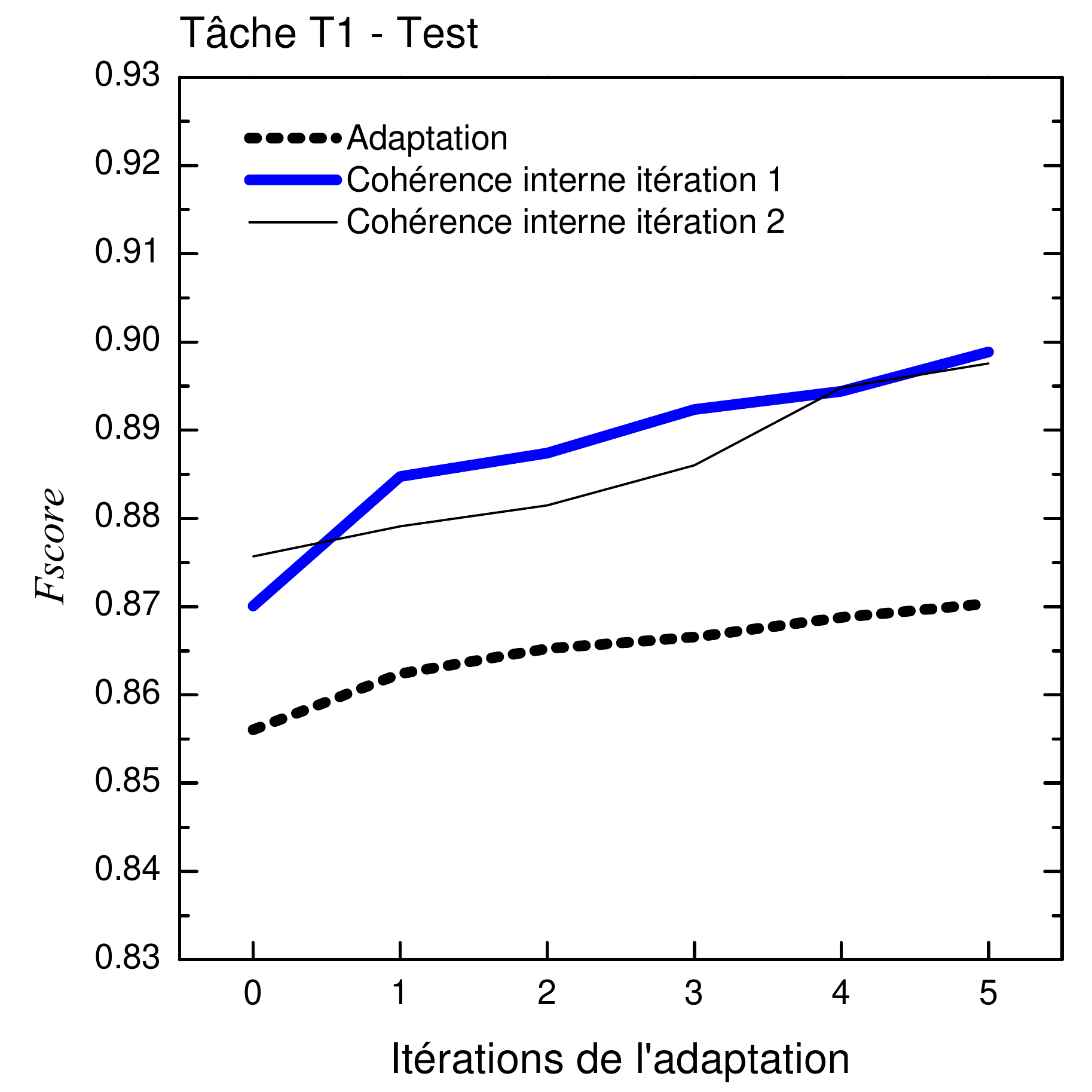}
 \caption{{\itshape Fscore} tâche T1 Modèle I / Adaptation vs Cohérence interne / corpus D et T.}
 \label{fig:tache1I}
\end{center}
\end{figure}

\begin{figure}[ht]
\begin{center}
 \includegraphics[width=6cm]{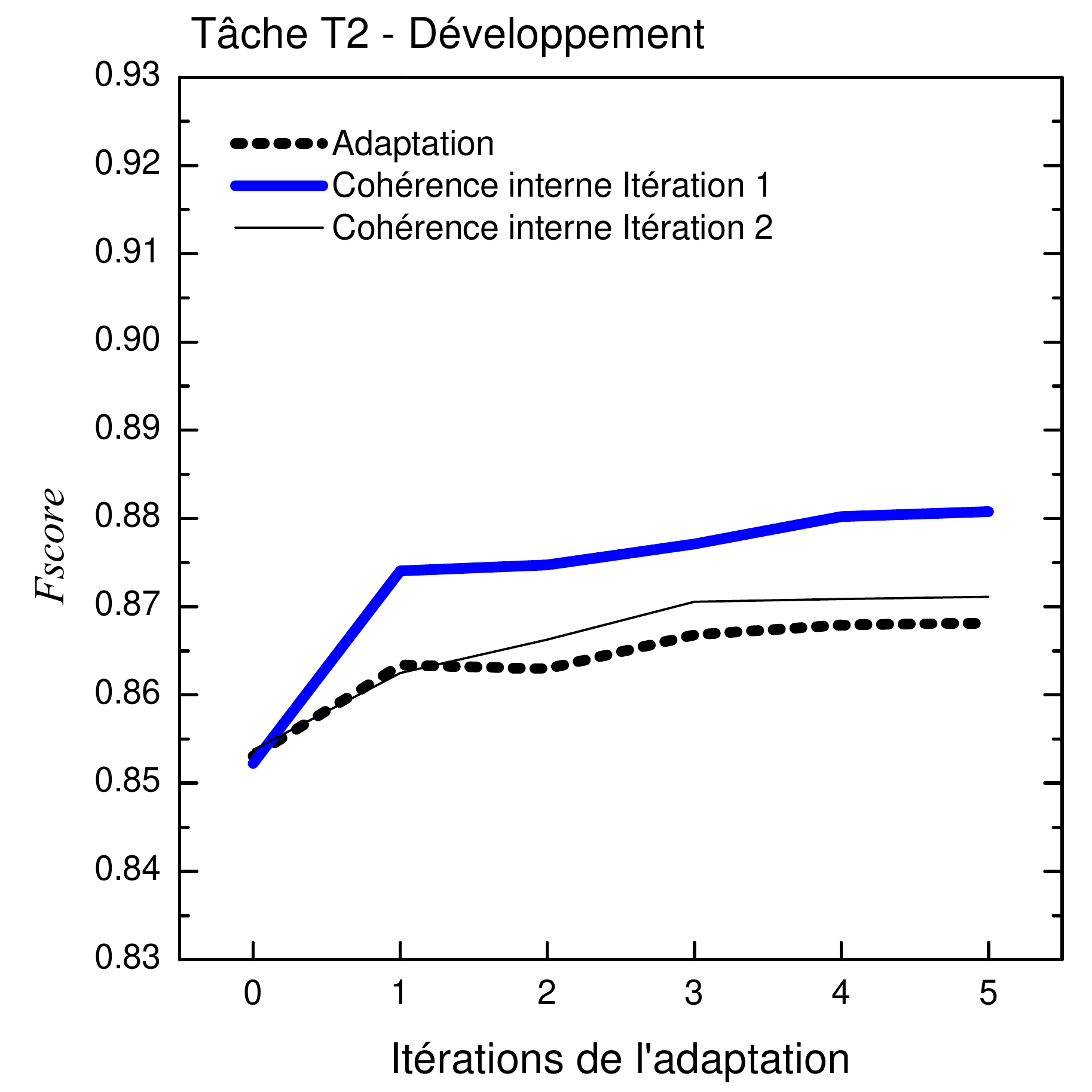}
 \includegraphics[width=6cm]{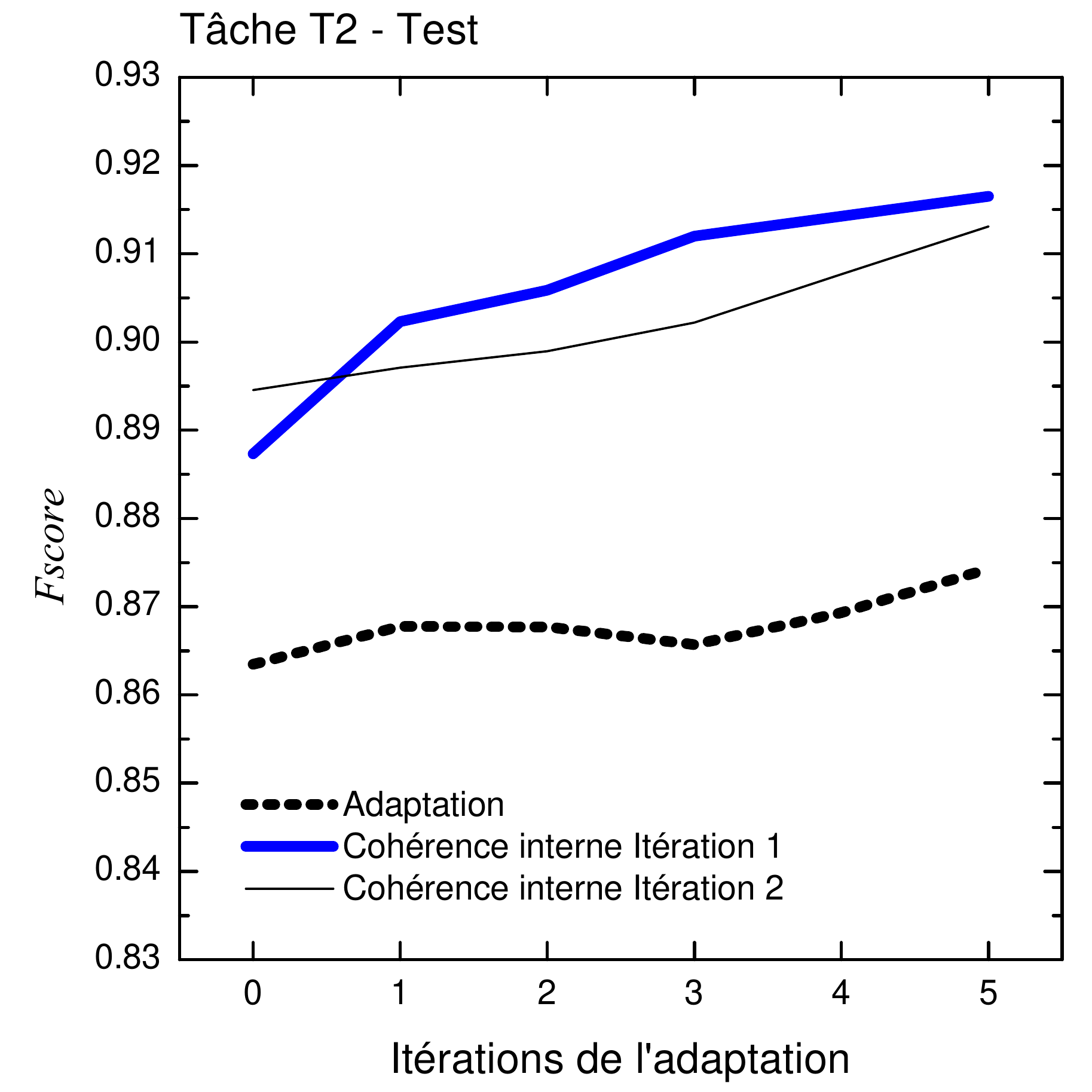}
 \caption{{\itshape Fscore} tâche T2 Modèle I / Adaptation vs Cohérence interne / corpus D et T.}
 \label{fig:tache2I}
\end{center}
\end{figure}

\begin{figure}[ht]
\begin{center}
 \includegraphics[width=6cm]{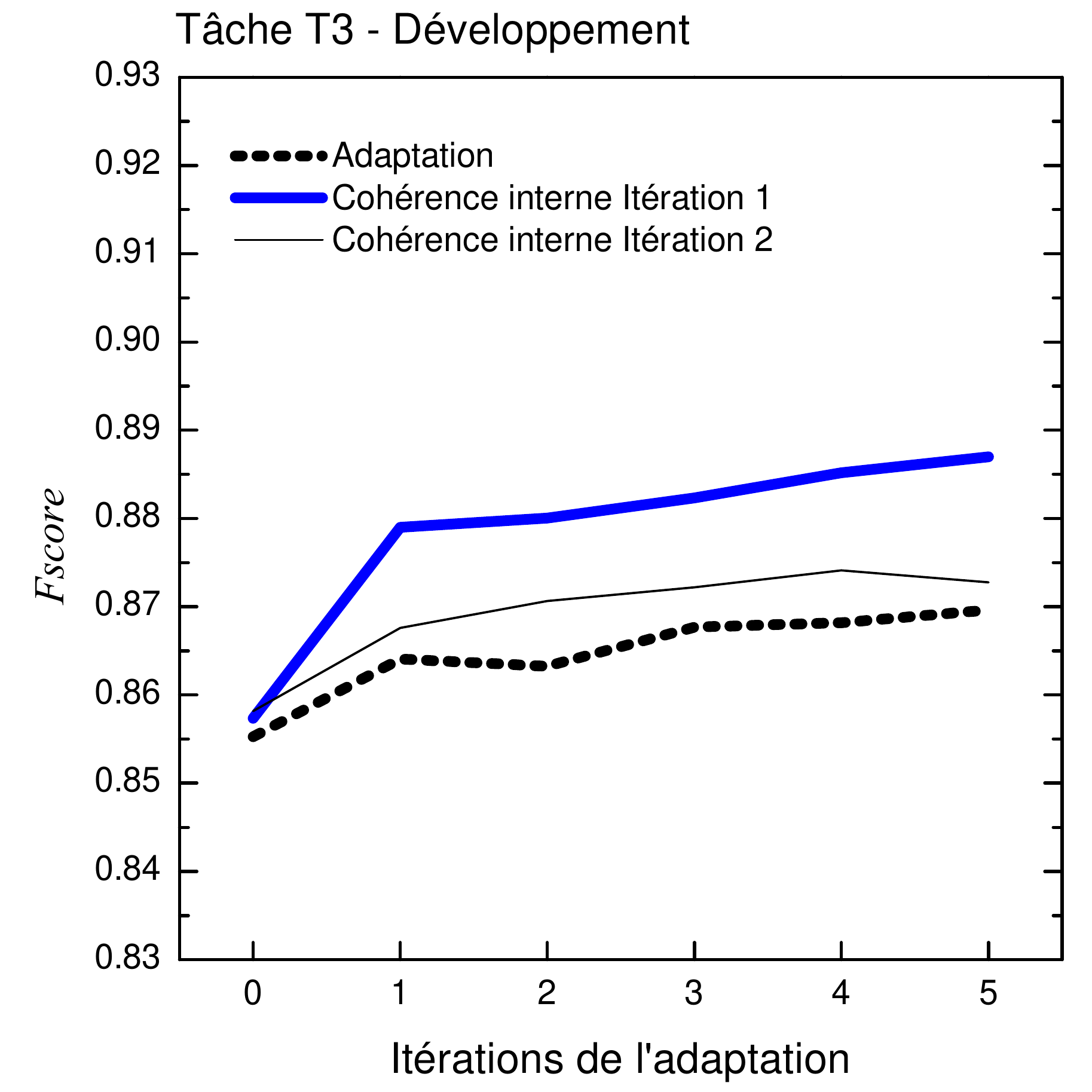}
 \includegraphics[width=6cm]{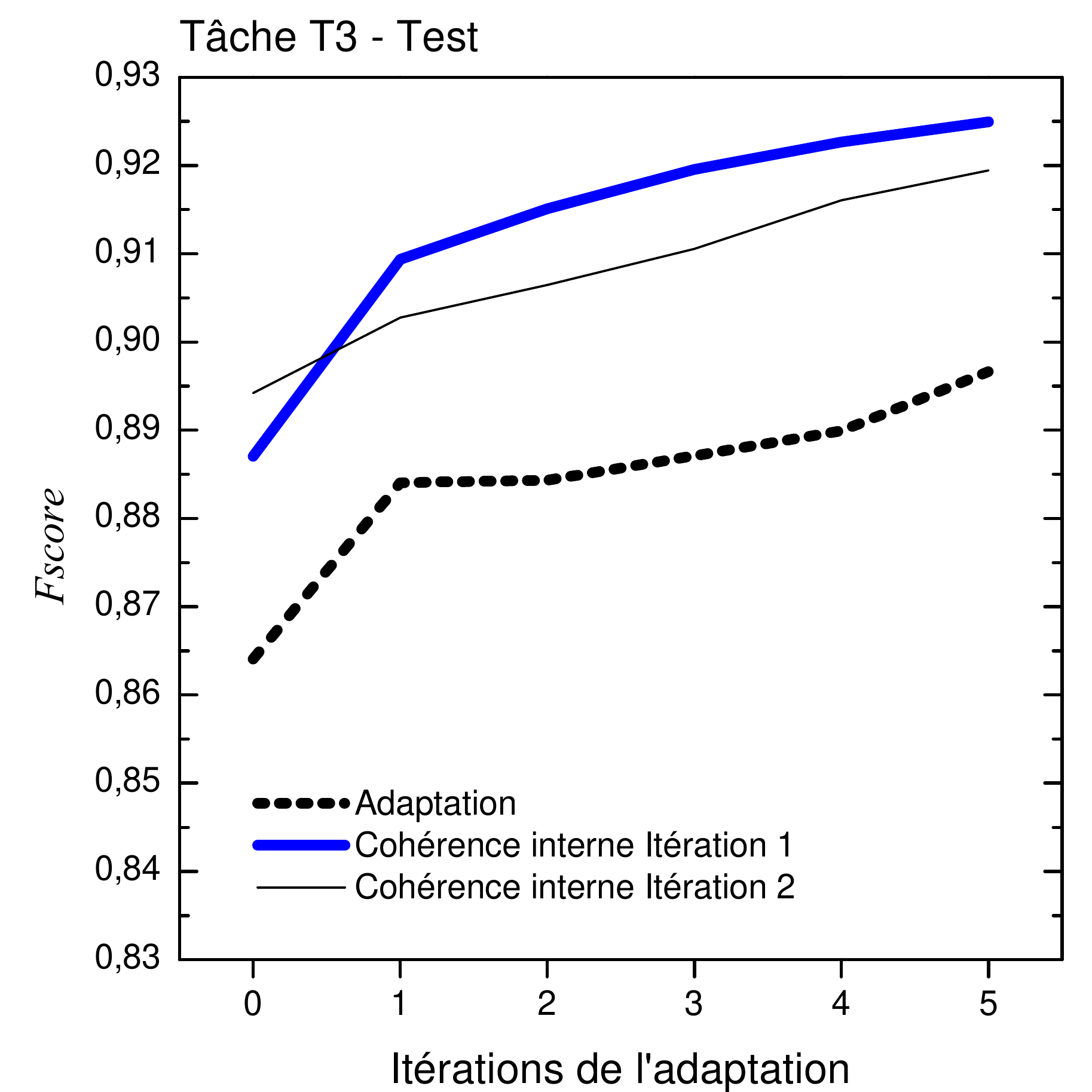}
 \caption{{\itshape Fscore}  tâche T3 Modèle I / Adaptation vs Cohérence interne / corpus D et T.}
 \label{fig:tache3I}
\end{center}
\end{figure}

Les résultats ont été améliorés grâce à la recherche d'une cohérence interne des discours. Cette étape intervient après application de l'automate markovien et avant la phase d'adaptation. Nous montrons, sur les figures \ref{fig:tache1I}, \ref{fig:tache2I} et \ref{fig:tache3I}, le {\itshape Fscore} obtenu pour les trois tâches à l'aide d'une adaptation plus la cohérence interne pour les corpus de Développement (D) et de Test (T). Dans tous les cas, l'axe horizontal représente les itérations de l'adaptation. Sur les graphiques, la ligne pointillée correspond aux valeurs du {\itshape Fscore} obtenues avec l'adaptation seule et les lignes continues à celles de la cohérence interne (une itération : ligne grosse ; deux itérations : ligne fine). Pour les trois tâches, on observe une amélioration notable du modèle de cohérence interne par rapport à celui de l'adaptation seule. Enfin, la valeur la plus élevée de {\itshape Fscore} est à présent obtenue pour la tâche T3 (figure \ref{fig:tache3I}), à un niveau de 0,925. Ce score dépasse largement le meilleur résultat ({\itshape Fscore} = 0,88) atteint lors du défi DEFT'05. Les valeurs précises des courbes sont rapportées dans les tableaux \ref{tab:modeleI} et \ref{tab:test1I}.

\begin{table}[ht]
 \begin{center}
   \tabcolsep = .6\tabcolsep
   \begin{tabular}{ccccccccc}
   \hline\hline
       \multicolumn{3}{c}{Tâche T1 (D)} &\multicolumn{3}{c}{Tâche T2 (D)}& \multicolumn{3}{c}{Tâche T3 (D)}\\
   \hline
   It & Adaptation & Cohérence puis & Adaptation & Cohérence puis & Adaptation & Cohérence puis\\ 
      & seule      & adaptation     & seule      & adaptation     & seule      & adaptation\\ 
   \hline
    0 & 0,841 & 0,850 & 0,853 & 0,852 & 0,855 & 0,857 \\
    1 & 0,845 & 0,871 & 0,863 & 0,874 & 0,864 & 0,879 \\
    2 & 0,853 & 0,872 & 0,863 & 0,875 & 0,863 & 0,880 \\
    3 & 0,857 & 0,875 & 0,867 & 0,877 & 0,868 & 0,882 \\
    4 & 0,860 & 0,877 & 0,868 & 0,880 & 0,868 & 0,885 \\
    5 & \textbf{0,862} & \textbf{0,877} & \textbf{0,868} & \textbf{0,881} & \textbf{0,870} & \textbf{0,887} \\
   \hline
   \end{tabular}
 \caption{Modèle I {\itshape Fscore} Développement : Adaptation seule et Cohérence interne.}
 \label{tab:modeleI}
 \end{center}
\end{table}

\begin{table}[ht]
 \begin{center}
   \tabcolsep = .6\tabcolsep
   \begin{tabular}{ccccccccc}
   \hline\hline
       \multicolumn{3}{c}{Tâche T1 (T)} &\multicolumn{3}{c}{Tâche T2 (T)} &\multicolumn{3}{c}{Tâche T3 (T)}\\
   \hline
   It & Adaptation & Cohérence puis & Adaptation & Cohérence puis & Adaptation & Cohérence puis\\ 
      & seule      & adaptation     & seule      & adaptation     & seule      & adaptation\\ 
\hline
    0 & 0,856 & 0,870 & 0,863 & 0,887 & 0,864 & 0,887 \\
    1 & 0,862 & 0,885 & 0,868 & 0,902 & 0,884 & 0,909 \\
    2 & 0,865 & 0,888 & 0,868 & 0,906 & 0,884 & 0,915 \\
    3 & 0,867 & 0,892 & 0,866 & 0,912 & 0,887 & 0,920 \\
    4 & 0,869 & 0,894 & 0,869 & 0,914 & 0,890 & 0,923 \\
    5 & \textbf{0,870} & \textbf{0,899} & \textbf{0,874} & \textbf{0,917} & \textbf{0,897} & \textbf{0,925} \\
   \hline
   \end{tabular}
 \caption{Modèle I $Fscore$ Test : Adaptation seule et Cohérence interne.}
 \label{tab:test1I}
 \end{center}
\end{table}

\begin{figure}[ht]
\begin{center}
 \includegraphics[width=6cm]{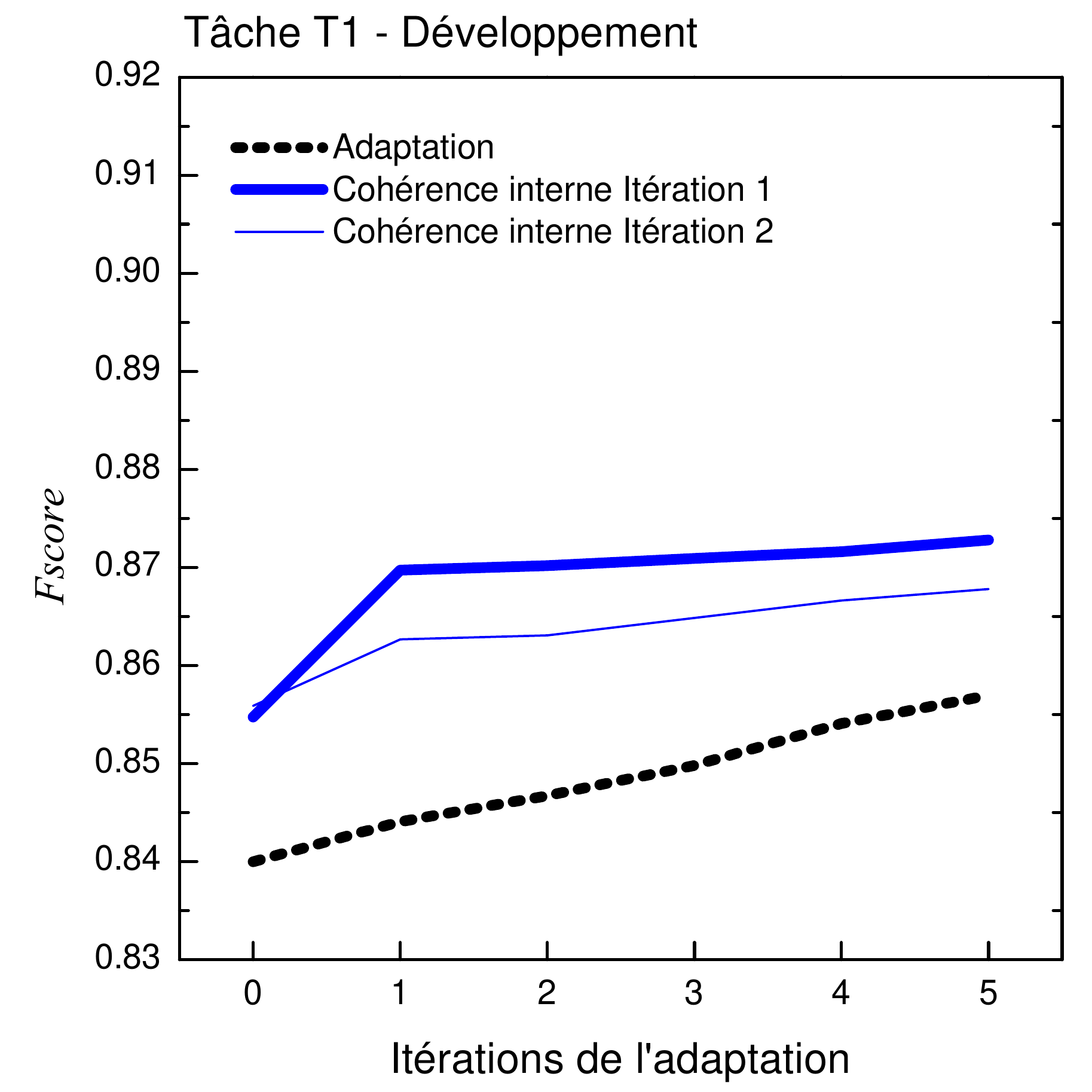}
 \includegraphics[width=6cm]{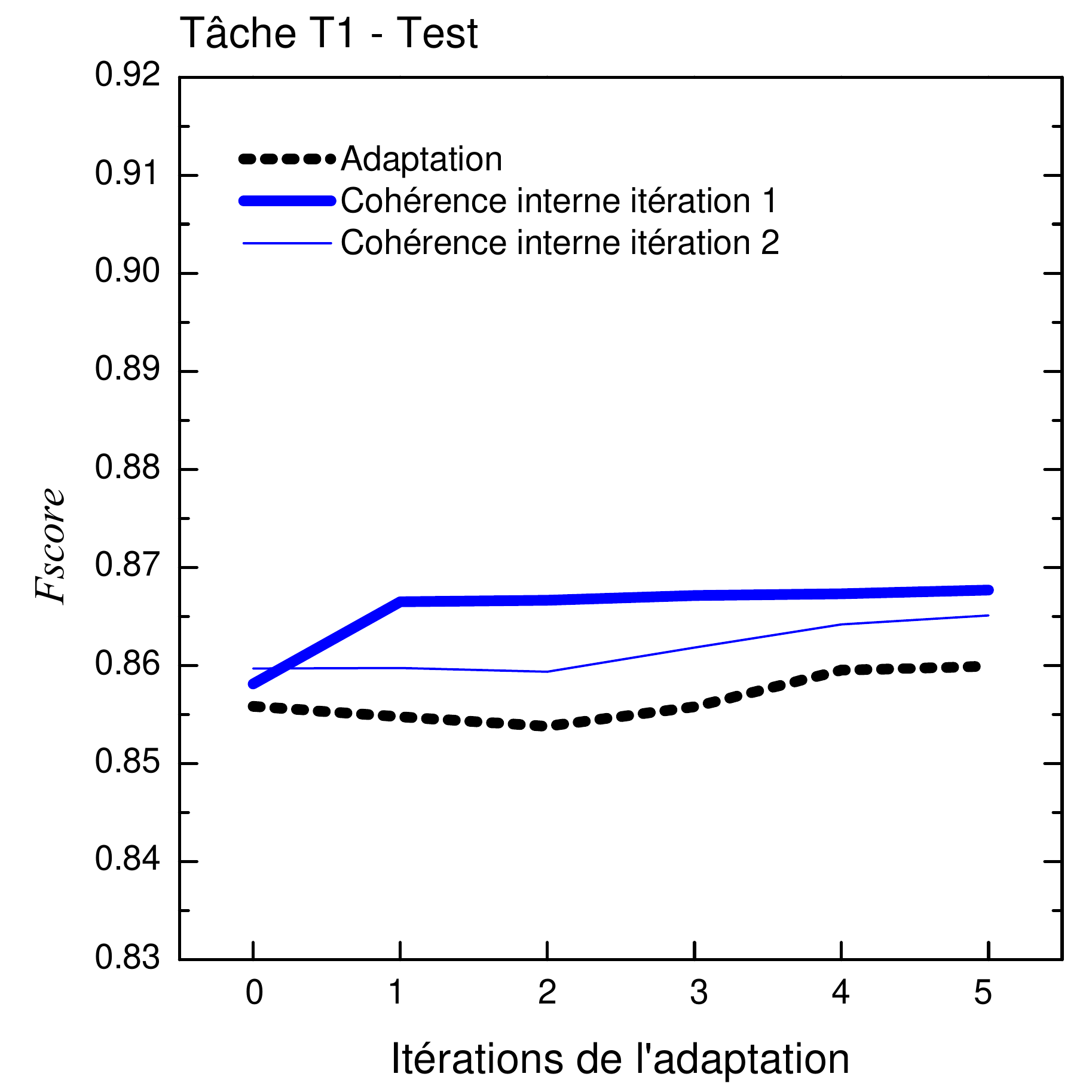}
 \caption{{\itshape Fscore} tâche T1 Modèle II / Adaptation vs Cohérence interne / corpus D et T.}
 \label{fig:tache1II}
\end{center}
\end{figure}

\begin{figure}[ht]
\begin{center}
 \includegraphics[width=6cm]{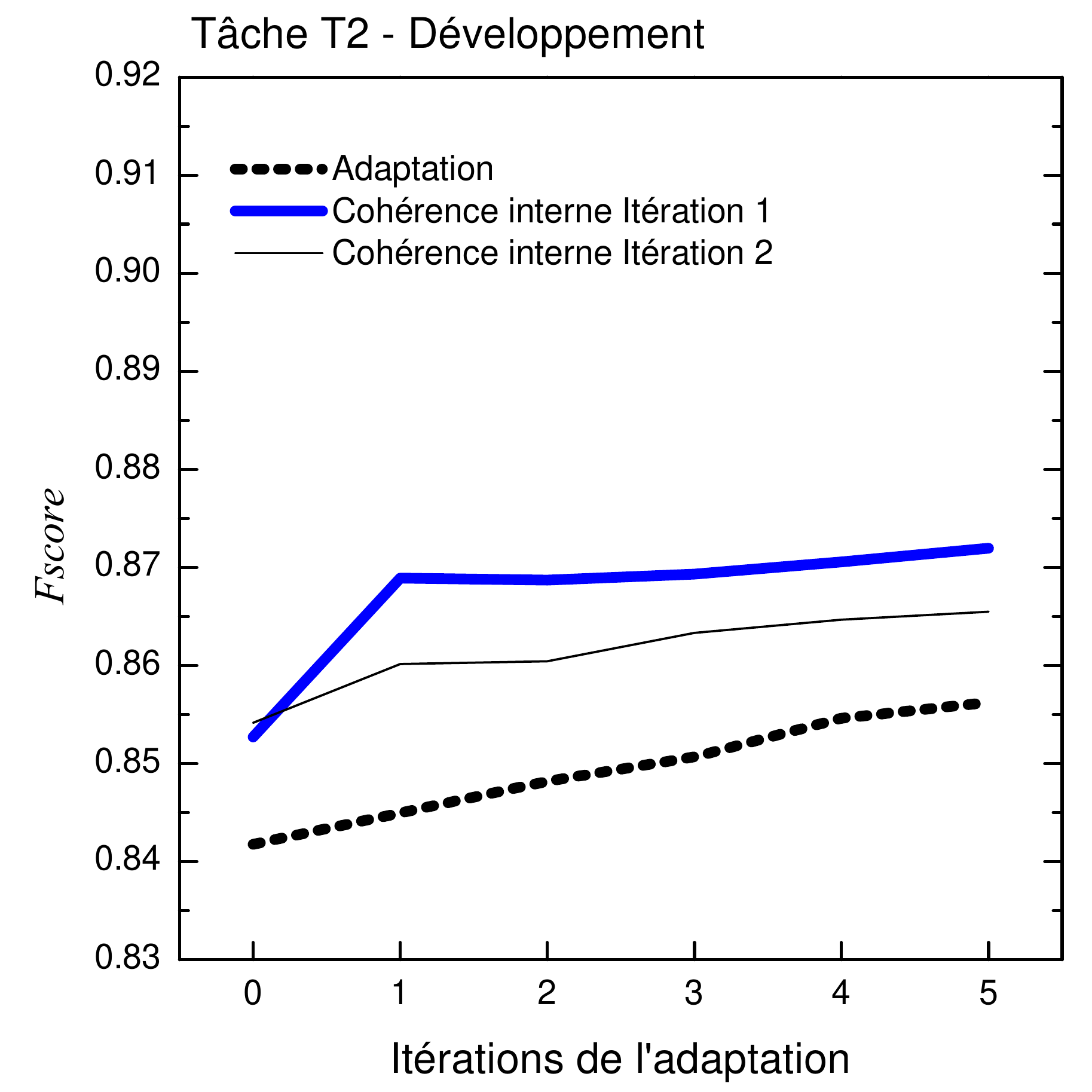}
 \includegraphics[width=6cm]{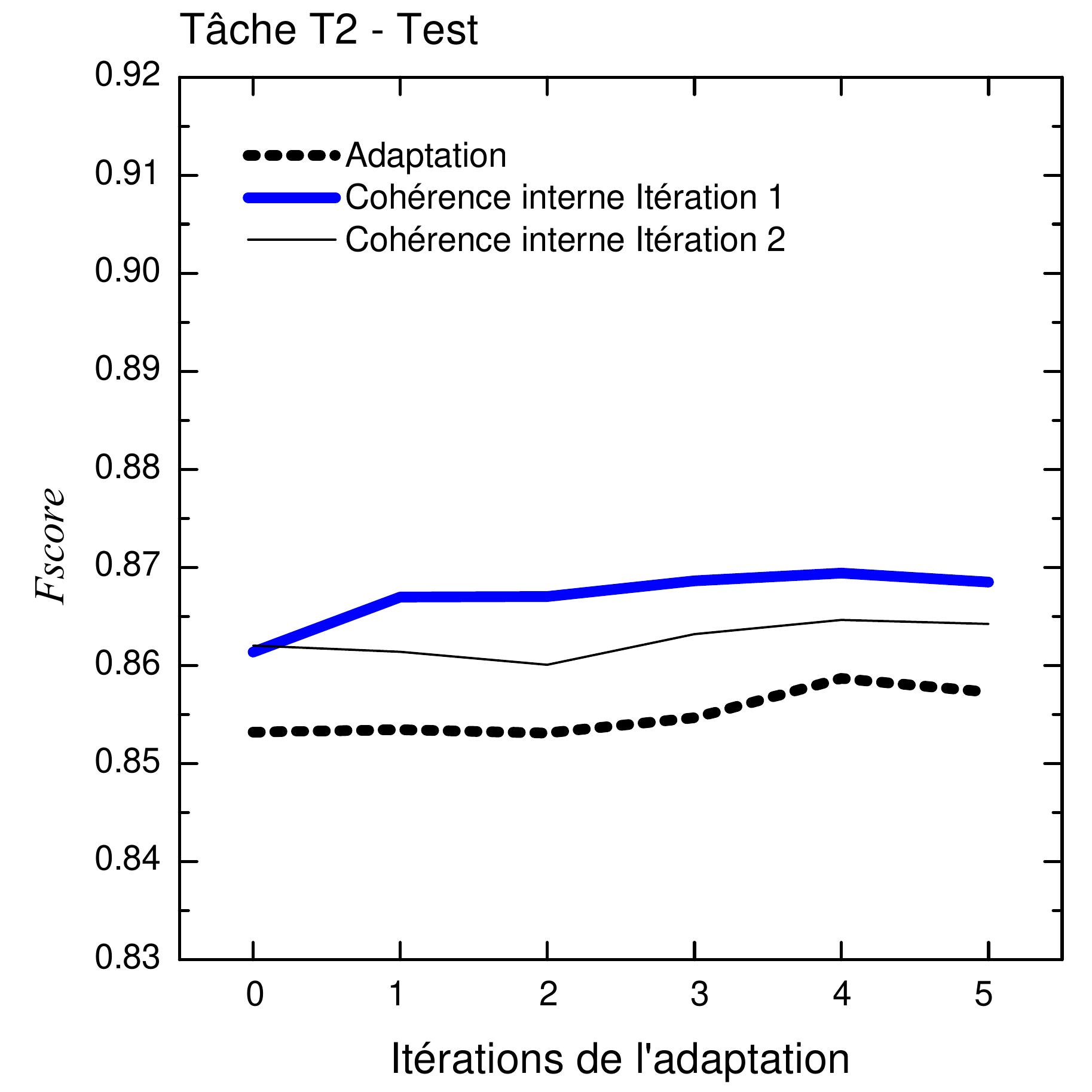}
 \caption{{\itshape Fscore} tâche T2 Modèle II / Adaptation vs Cohérence interne / corpus D et T.}
 \label{fig:tache2II}
\end{center}
\end{figure}

\begin{figure}[ht]
\begin{center}
 \includegraphics[width=6cm]{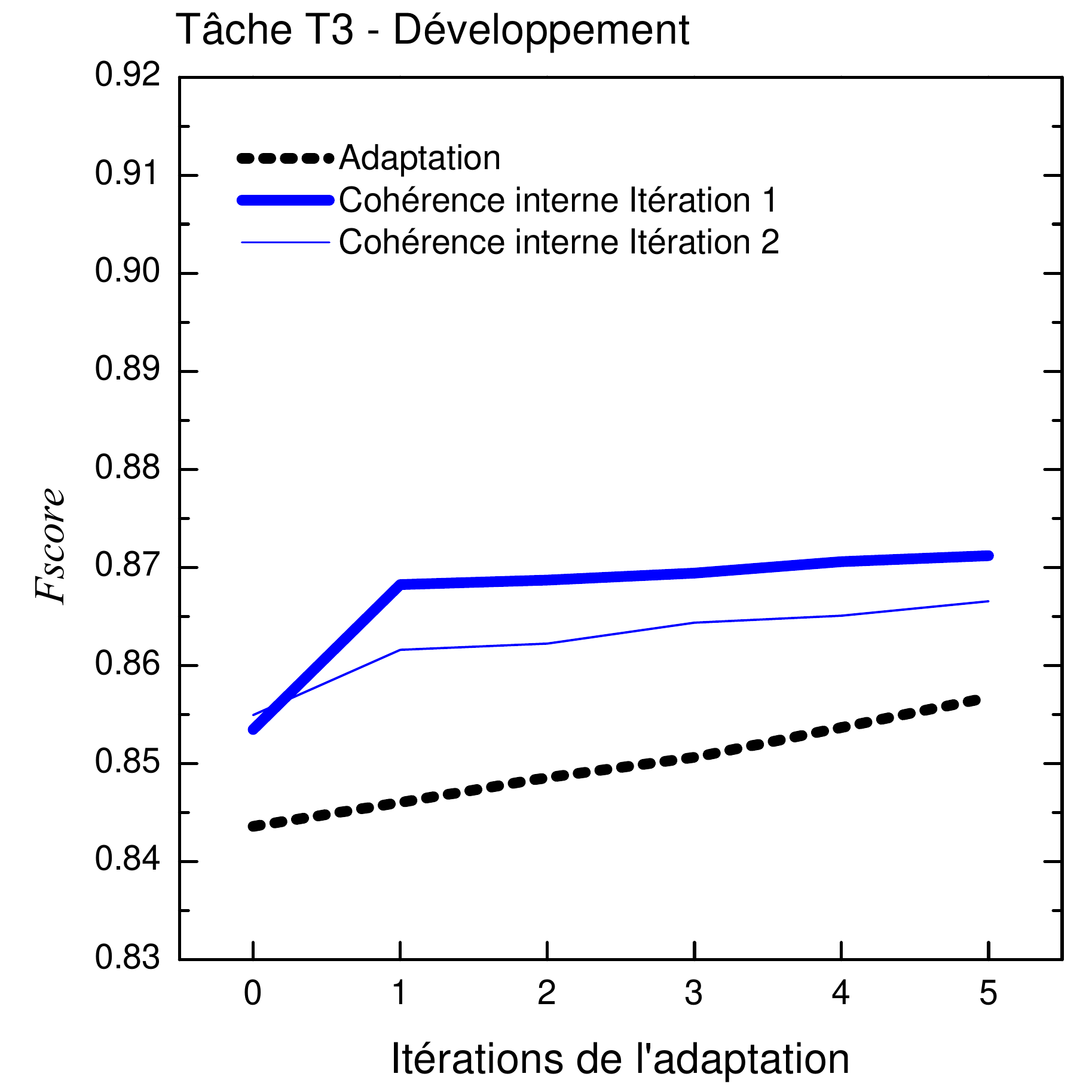}
 \includegraphics[width=6cm]{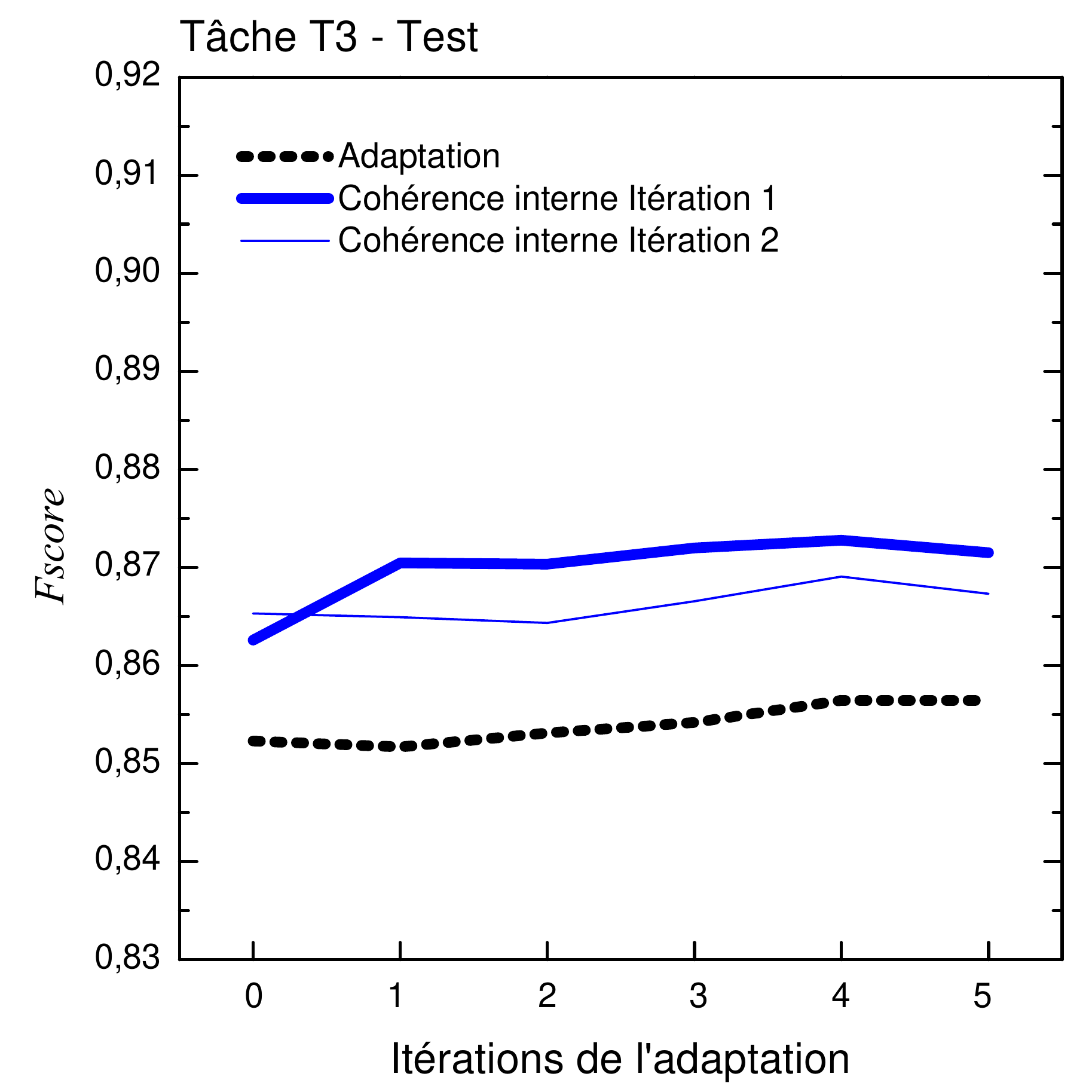}
 \caption{{\itshape Fscore}  tâche T3 Modèle II / Adaptation vs Cohérence interne / corpus D et T.}
 \label{fig:tache3II}
\end{center}
\end{figure}

Les figures \ref{fig:tache1II}, \ref{fig:tache2II} et \ref{fig:tache3II} montrent, avec la même convention que les figures précédentes, le {\itshape Fscore} pour le modèle II, où n'ont été appliqués ni filtrage ni lemmatisation. Ici encore, les valeurs les plus élevées sont obtenues pour la tâche T3, avec un {\itshape Fscore} = 0,873. La comparaison avec les performances ({\itshape Fscore} = 0,801 pour la tâche T3) de ce même modèle que nous avons employé lors du défi DEFT'05, est avantageuse : sept points de plus\footnote{Plus de détails sont rapportés aux tableaux \ref{tab:modeleII} et \ref{tab:testII}.}. Cette amélioration est due essentiellement à la cohérence interne et permet d'approcher la meilleure valeur ({\itshape Fscore} = 0,881 rapporté dans \cite{el-beze:2005}) qui avait été obtenue avec l'adaptation seule et un filtrage et lemmatisation préalables. Bien que l'utilisation de ce modèle soit un peu moins performante (et de ce fait contestée), nous pensons qu'il peut être utile d'y recourir, si l'on veut éviter la lourdeur des certains processus de prétraitement.

\begin{table}[ht]
 \begin{center}
   \tabcolsep = .6\tabcolsep
   \begin{tabular}{ccccccccc}
   \hline\hline
       \multicolumn{3}{c}{Tâche T1 (D)} &\multicolumn{3}{c}{Tâche T2 (D)} &\multicolumn{3}{c}{Tâche T3 (D)}\\
   \hline
   It & Adaptation & Cohérence puis & Adaptation & Cohérence puis & Adaptation & Cohérence puis\\ 
      & seule      & adaptation     & seule      & adaptation     & seule      & adaptation\\ 
\hline
    0 & 0,834 & 0,867 & 0,842 & 0,868 & 0,844 & 0,867 \\
    1 & 0,844 & 0,881 & 0,845 & 0,881 & 0,846 & 0,881 \\
    2 & 0,847 & 0,881 & 0,848 & 0,882 & 0,849 & 0,883 \\
    3 & 0,850 & 0,881 & 0,851 & 0,882 & 0,851 & 0,883 \\
    4 & 0,854 & 0,881 & 0,855 & \textbf{0,883} & 0,854 & \textbf{0,883} \\
    5 & \textbf{0,857} & \textbf{0,882} & \textbf{0,856} & 0,882 & \textbf{0,857} & 0,882 \\
   \hline
   \end{tabular}
 \caption{Modèle II $Fscore$ Développement : Adaptation seule et Cohérence interne.}
 \label{tab:modeleII}
 \end{center}
\end{table}

\begin{table}[ht]
 \begin{center}
   \tabcolsep = .6\tabcolsep
   \begin{tabular}{ccccccccc}
   \hline\hline  
       \multicolumn{3}{c}{Tâche T1 (T)} &\multicolumn{3}{c}{Tâche T2 (T)}& \multicolumn{3}{c}{Tâche T3 (T)}\\
   \hline
   It & Adaptation & Cohérence puis & Adaptation & Cohérence puis & Adaptation & Cohérence puis\\ 
      & seule      & adaptation     & seule      & adaptation     & seule      & adaptation\\ 
\hline
    0 & 0,856 & 0,857 & 0,853 & 0,860 & 0,852 & 0,859 \\
    1 & 0,855 & 0,873 & 0,853 & 0,877 & 0,852 & 0,872 \\
    2 & 0,854 & 0,873 & 0,853 & 0,877 & 0,853 & 0,873 \\
    3 & 0,856 & \textbf{0,873} & 0,855 & 0,877 & 0,854 & 0,874 \\
    4 & 0,860 & 0,872 & \textbf{0,859} & 0,877 & 0,856 & 0,874 \\
    5 & \textbf{0,860} & 0,871 & 0,858 & \textbf{0,877} & \textbf{0,856} & \textbf{0,874} \\
   \hline
   \end{tabular}
 \caption{Modèle II $Fscore$ Test : Adaptation seule et Cohérence interne.}
 \label{tab:testII}
 \end{center}
\end{table}

\subsection{Fusion de méthodes}

Le dernier test que nous avons réalisé fait appel à une fusion de l'ensemble des modèles. Nous avons appliqué un algorithme de vote sur presque toutes les hypothèses issues des modèles I et II. Les hypothèses (qui vont faire office de juges) proviennent des différentes itérations de l'adaptation, avec ou sans cohérence interne. Nous avons tenu compte des avis d'un nombre de juges donné ({\itshape NbJ}), en pondérant l'avis de chaque juge {\itshape j} par un poids $\alpha _{j}$ de telle sorte que le critère de décision final est le suivant :
\begin{equation}
  \theta_i = signe \left( \sum^{NbJ}_{j=1} \alpha_{j} \xi_{i,j} - \delta \right)
\end{equation}

\noindent Si $\theta_i$ est négatif alors l'étiquette du segment $i$ sera {\itshape C} ; {\itshape M} autrement.
Avec $\alpha_j \in \Re$, $\beta_{i,j} \in \{0,1\}$ et la convention 0 = $C$ et 1 = $M$.

La stratégie est la suivante : afin d'avoir un degré de confiance suffisant, il faut retenir les segments auxquels une majorité de juges attribue l'étiquette $M$. Les paramètres $\alpha_j$ et $\delta$ ont été ajustés pour minimiser le nombre d'erreurs sur l'ensemble de développement (D). Nous nous sommes proposés de voir cette estimation comme un problème de classification à $NbJ$ entrées et une sortie, c'est-à-dire, comme un problème d'apprentissage supervisé. Nous avons ainsi défini un exemple d'apprentissage comme le vecteur binaire $\xi_j = \{0,1\}, j=1,\cdots,NbJ$. La sortie (classe de référence) de cet exemple est un scalaire $\tau=\{-1,1\}$ ($-1$ pour la classe $C$, $+1$ pour la classe $M$). L'ensemble d'apprentissage est donc constitué de $S$ segments et $NbJ$ juges, et nous le dénoterons par $\aleph = \{\vec\xi_i,\tau_i\}; i=1,\cdots,S$. Trouver les poids $\alpha_j$ correspond donc à trouver les $j$ poids d'un perceptron entraîné sur l'ensemble $\aleph$. Nous avons utilisé un perceptron optimal à recuit déterministe\footnote{La position de l'hyperplan séparateur des classes se fait par une modification progressive des poids (descente en gradient) contrôlés au moyen d'une température de recuit lors de l'apprentissage.} entraîné par l'algorithme Minimerror \cite{Minimerror,Monoplane,torres-moreno:2002}, où l'apprentissage garantit que si l'ensemble $\aleph$ est linéairement séparable, l'algorithme trouve la solution optimale (marge maximale de séparation) et s'il ne l'est pas (comme cela semble être le cas ici), il trouve une solution qui minimise le nombre de fautes commises. Ainsi, nous avons trouvé un seuil $\delta$ = 8,834 et les poids $\alpha_j$ avec un nombre de juges $NbJ$ = 89. 

Pour l'ensemble de développement, les résultats obtenus au moyen de cette fusion sont encore meilleurs qu'avec les autres méthodes. Nous obtenons, dans ce cas, un {\itshape Fscore} = 0,914 avec une précision de 0,916 et un rappel de 0,911. Cependant, pour l'ensemble de test, la fusion ne dépasse pas le meilleur résultat obtenu jusqu'à présent. En effet, on atteint un {\itshape Fscore} = 0,914, avec une précision de 0,892 et un rappel de 0,937. Il est connu que les perceptrons (et les réseaux de neurones en général) trouvent parfois des valeurs de poids trop bien adaptées à l'ensemble d'apprentissage (phénomène de sur-apprentissage). Le fait de n'avoir pas eu de meilleurs résultats sur le test le confirme. Cependant, nous pensons que si les $\vec\xi_{i}$ étaient des probabilités au lieu d'être des 0 et des 1, on aurait pu observer un meilleur comportement.

\subsection{Analyse des erreurs}

Nous avons analysé les erreurs commises par notre système. Sur un total de 27 163 phrases de l'ensemble de Test de la tâche T3, le Modèle I avec la méthode d'adaptation et la cohérence interne des discours, a fait un total de 578 erreurs ({\itshape Fscore} = 0,925) :
\begin{itemize}
    \item 233 erreurs de la classe {\itshape C}  (faux négatifs assimilés au rappel), dont :
    \begin{itemize}
        \item 37 phrases {\itshape C} à la frontière inversée ($\approx$ 16\%) ;
        \item 113 phrases {\itshape C} en blocs ($\approx$ 49\%) ;
        \item 83 phrases {\itshape C} entre blocs {\itshape C} ($\approx$ 36\%) ;
    \end{itemize}
    \item 345 erreurs de la classe {\itshape M} (faux positifs assimilés à la précision), dont :
    \begin{itemize}
        \item 35 phrases {\itshape M} à la frontière inversée ($\approx$ 10\%) ;
        \item 126 phrases {\itshape M} en blocs ($\approx$ 37\%) ;
        \item 184 phrases {\itshape M} insérées dans 21 discours de classe {\itshape C} exclusive ($\approx$ 53\%).
    \end{itemize}
\end{itemize}

Le problème le plus grave concerne la précision (59\% du total des erreurs), et ici, la plus grande majorité (53\% de faux positifs) est due aux insertions des phrases {\itshape M} dans des discours de classe {\itshape
C}\footnote{Voir en annexe l'analyse du discours 520, concernant cette situation problématique.}. L'autre
problème se présente dans les 126 phrases en blocs inversés (37\%). Ces problèmes sont peut-être dus à
l'utilisation de la cohérence interne : sur le tableau \ref{tab:precisionT3MI}, on voit qu'en adaptation seule,
la précision est toujours plus élevée que le rappel (en D comme en T). Pour la cohérence interne, la situation
est inversée : le rappel est bien meilleur que la précision. Le même comportement a été retrouvé dans le Modèle
II. Un autre pourcentage important d'erreurs (49\% de faux positifs) a lieu dans l'inversion d'un nombre
important de blocs (113 phrases). Enfin, une autre partie non négligeable (10\% de faux positifs, 16\% de faux
négatifs) correspond à l'inversion de catégorie d'une phrase unique à la frontière des découpages (soit $i$ ou
$j$, voir figure \ref{fig:decoupage}). La détection de cette frontière, reste un sujet très délicat avec nos approches.
\begin{table}[ht]
 \begin{center}
   \tabcolsep = 1.1\tabcolsep
   \begin{tabular}{ccccc}
   \hline\hline
      & \multicolumn{2}{c}{Adaptation} & \multicolumn{2}{c}{Cohérence interne}\\
      & \multicolumn{2}{c}{} & \multicolumn{2}{c}{puis adaptation}\\
   \hline      
                 & Précision & Rappel & Précision & Rappel \\
   \hline
   Développement & \textbf{0,918}     & 0,826  & 0,882     & \textbf{0,892} \\
   \hline
   Test          & \textbf{0,931}     & 0,866  & 0,912     & \textbf{0,939} \\
   \hline
   \end{tabular}
 \caption{Précision et Rappel pour la tâche T3, Modèle I, adaptation et cohérence interne à la dernière  
          itération de l'adaptation.}
 \label{tab:precisionT3MI}
 \end{center}
\end{table}

\begin{figure}[ht]
\begin{center}
 \includegraphics[width=5.7cm]{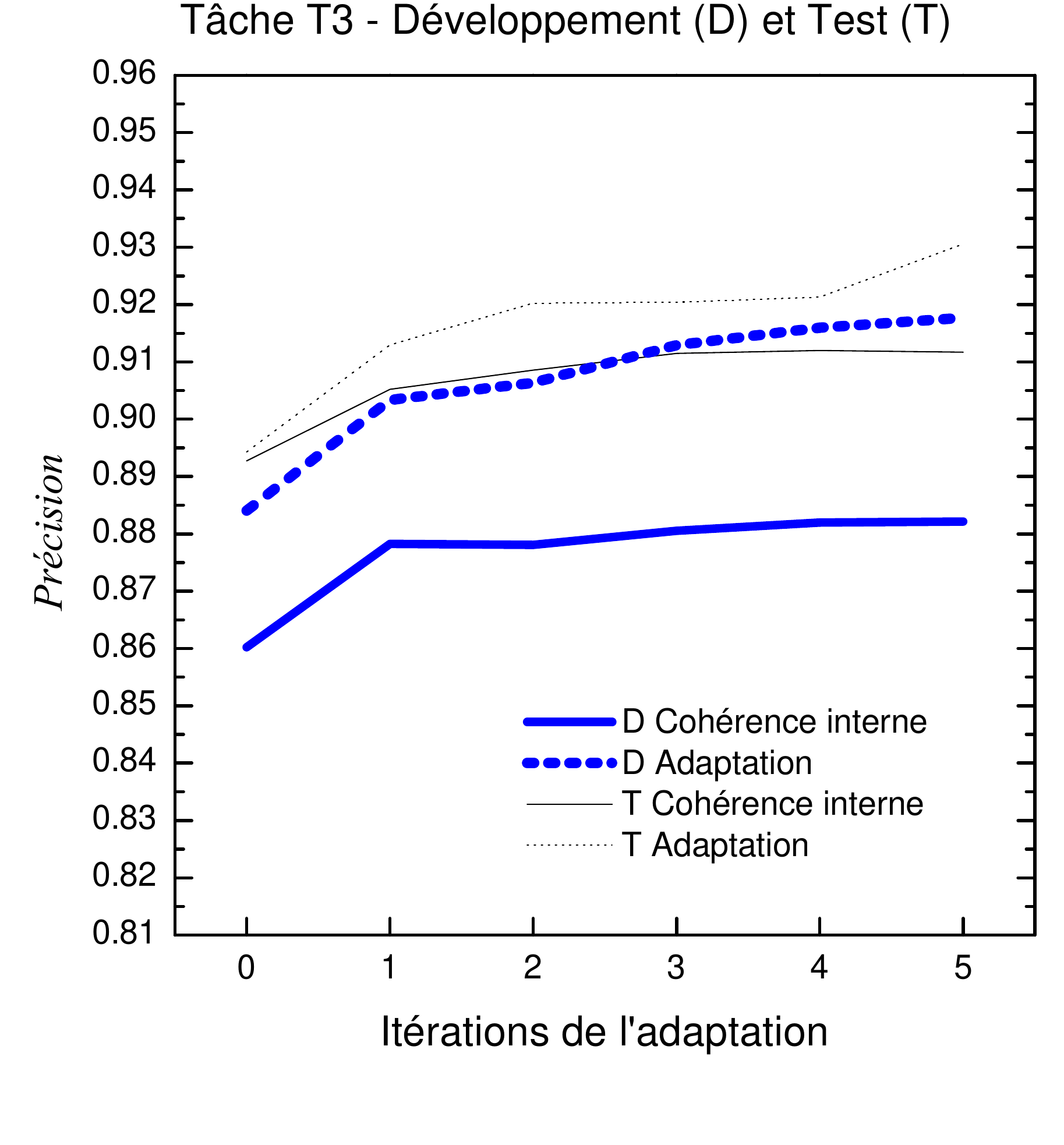}
 \includegraphics[width=5.7cm]{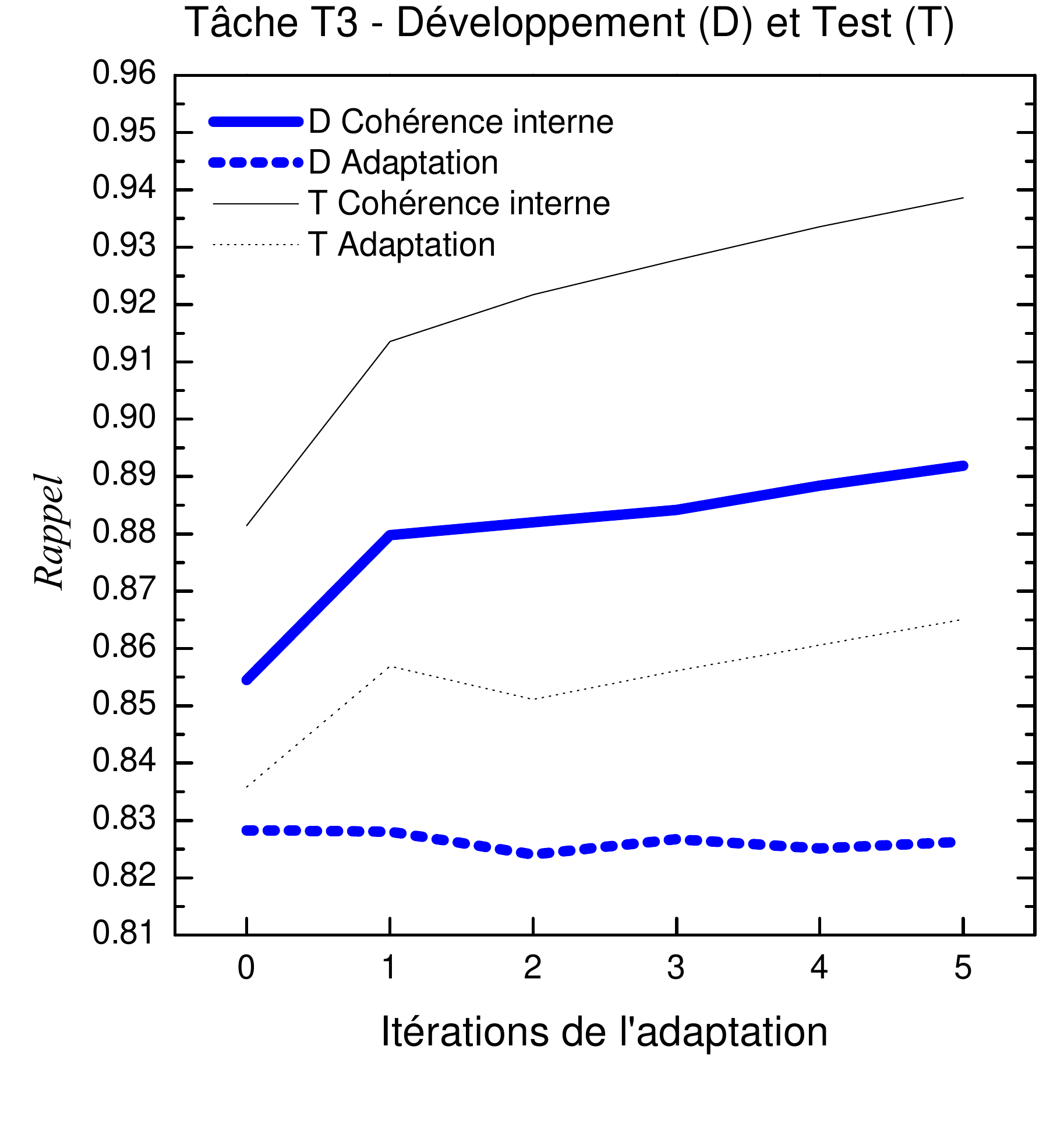}
 \caption{Précision-Rappel Modèle I / tâche T3 : Adaptation vs Cohérence interne.}
 \label{fig:precisionM3T3}
\end{center}
\end{figure}

\begin{figure}[ht]
\begin{center}
 \includegraphics[width=10cm]{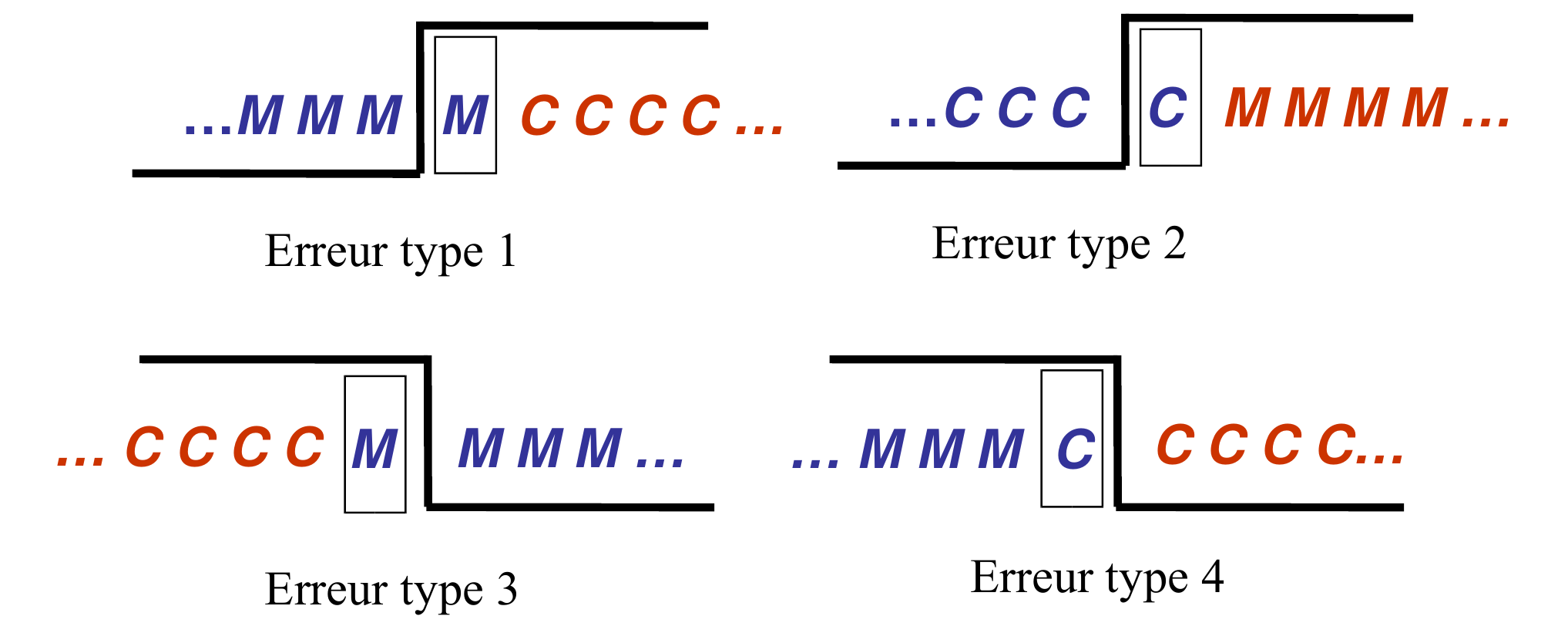}
 \caption{Types d'erreurs frontière de bloc.}
 \label{fig:erreurs}
\end{center}
\end{figure}

La figure \ref{fig:precisionM3T3} montre les courbes de Précision et de Rappel. Pour rester concis, nous allons
présenter ici seulement les résultats correspondant à la tâche T3 du modèle I. La cohérence interne est affichée
uniquement sur la première itération. Dans les deux cas, nous montrons des résultats sur les corpus de
Développement (D) et Test (T). On voit que sur l'ensemble de test T, la précision de la cohérence interne puis
adaptation est moins élevée que celle de l'adaptation seule. La même situation se produit pour le rappel.
Néanmoins, un phénomène d'inversion se présente en développement : en rappel on est plus performant avec la
cohérence qu'avec l'adaptation seule, et en précision avec l'adaptation seule que avec la cohérence.
Nous avons aussi calculé la longueur moyenne (en mots) des segments mal classés (tâche T3 / modèle I)
suivant le type d'erreur (cf. figure \ref{fig:erreurs}) : discours exclusifs de la classe {\itshape C}, longueur
moyenne $\approx$ 21. Erreurs de début du bloc : type 1 longueur moyenne $\approx$ 21 ; type 2 longueur moyenne
$\approx$ 22. Erreurs de fin du bloc : type 3 longueur moyenne $\approx$ 21 ; type 4 longueur moyenne $\approx$
26. Cependant, il est difficile de tirer de conclusions à partir de cette information.
Enfin, de façon comparative les résultats du {\itshape Fscore} sur les trois tâches sans adaptation, montrés au
tableau \ref{tab:fscoreSA}, confirment l'importance de l'utilisation de l'automate de Markov : il fait un gain
global de $\approx$ 25\% avec les deux modèles proposés.
\begin{table}[ht]
 \begin{center}
   \tabcolsep = 1.1\tabcolsep
   \begin{tabular}{ccccccc}
   \hline\hline
      & \multicolumn{3}{c}{Développement} & \multicolumn{3}{c}{Test}\\
   \hline
       & T1 & T2 & T3 & T1 & T2 & T3 \\
	 \hline
    Modèle I & 0,570 & 0,570 & \textbf{0,574} & 0,593 & 0,595 & \textbf{0,596} \\
   \hline
    Modèle II& 0,549 & 0,551 & \textbf{0,555} & 0,581 & 0,582 & \textbf{0,585} \\
   \hline
   \end{tabular}
 \caption{{\itshape Fscore} pour les trois tâches sans adaptation.}
 \label{tab:fscoreSA}
 \end{center}
\end{table}

\section{Conclusions et perspectives}
\label{sec:conclusions}

Nous avons introduit une formalisation de la cohérence interne des discours qui a beaucoup amélioré nos
résultats rapportés en \cite{el-beze:2005}. Cette cohérence ainsi que l'adaptation ont été combinées
conjointement avec les modèles d'apprentissage préalablement développés, comme la modélisation bayésienne qui semble déterminant, l'automate de Markov et des processus d'adaptation. Les résultats que nous obtenons pour la tâche T3 avec la cohérence interne en terme de {\itshape Fscore} = 0,925 sont très encourageants. Cependant, l'utilisation de la cohérence interne présente un risque : quelques phrases avec une thématique différente, peuvent faire basculer tout un bloc vers l'autre étiquette. Ce type de comportement local entraîne des instabilités globales (semblables à ce qui se produit dans le jeu « Reversi »), dont la prévision reste très difficile, ayant comme conséquence une baisse générale des performances. Ne pas lemmatiser et ne rien filtrer dégrade un peu les performances ({\itshape Fscore} = 0,874 avec le modèle II) mais permet d'éviter l'application d'un processus additionnel de prétraitement qui pour certaines langues peut être relativement lourd. La fusion des hypothèses vue comme un vote de plusieurs juges pondérés par un perceptron optimal a permis de surpasser les résultats précédents en développement ({\itshape
Fscore} = 0,914). Cependant nous pensons qu'il reste encore du travail pour améliorer cette stratégie afin d'obtenir de meilleures performances en test.
Des études comme celle de \cite{rigouste:2005} sur le même corpus confirment que l'utilisation de méthodes probabilistes est la mieux adaptée à ce type de segmentation thématique. Le recours à un réseau de Noms Propres est utile et nous encourage par la suite à employer une ressource lexicale comme \cite{eurowordnet:1998} pour tirer parti de réseaux sur les noms communs. Pour s'affranchir des contraintes liées à la constitution d'une ressource sémantique, il serait judicieux de recourir à des approches telles que \emph{Latent Semantic Analysis}
\cite{deerwester:1990} ou PLSA \cite{hofmann:1999}. D'autres perspectives d'application, comme celle de la séparation de thèmes sont aussi envisageables. Il faut reconnaître, cependant, que s'il s'était agi de traiter un texte composite moins artificiel que celui proposé par DEFT, par exemple un dialogue, la difficulté aurait été accrue. Des frontières thématiques ne coïncident pas forcément avec des débuts de phrase. Les thèmes peuvent s'entremêler et composer un tissu discursif où les fils sont enchevêtrés de façon subtile. Beaucoup reste à faire pour pouvoir différencier plusieurs orateurs ou plusieurs thèmes comme envisagé dans le cadre du Projet
Carmel \cite{chen:2005}.

\bibliographystyle{plain}

\bibliography{biblio_rnti}

\begin{thebibliography}{10}

\bibitem{alphonse:2005}
E.~Alphonse, A.~Amrani, J.~Azé, T.~Heitz, A.-D Mezaour, and M.~Roche.
\newblock Préparation des données et analyse des résultats de {DEFT'05}.
\newblock In {\em Proc. of TALN 2005 - Atelier DEFT'05}, volume~2, pages
  95--97, 2005.

\bibitem{aze:2005}
J.~Azé and M.~Roche.
\newblock Présentation de l'atelier {DEFT'05}.
\newblock In {\em Proc. of TALN 2005 - Atelier {DEFT'05}}, volume~2, pages
  99--111, 2005.

\bibitem{bellot:2003}
P.~Bellot, E.~Crestan, M.~El-Bèze, L.~Gillard, and C.~De Loupy.
\newblock Coupling named entity recognition, vector-space model and knowledge
  bases for {TREC-11}, question-answering track.
\newblock In {\em Proceedings of TREC'02, Gaithersburg, USA, NIST Special
  publication 500 251}, 2003.

\bibitem{chen:2005}
B.~Chen, M.~El-Bèze, M.~Haddara, O.~Kraif, and G.~Moreau de~Montcheuil.
\newblock Contextes multilingues alignés pour la désambiguïsation
  sémantique : une étude expérimentale.
\newblock In {\em Proc. of Traitement Automatique des Langues Naturelles 2005},
  volume~1, pages 415--418, 2005.

\bibitem{collobert:2001}
R.~Collobert and S.~Bengio.
\newblock Support vector machines for large-scale regression problems.
\newblock {\em Journal of Machine Learning Research}, 1:143--160, 2001.

\bibitem{cotteret:1969}
J.-M. Cotteret and R.~Moreau.
\newblock {\em Le vocabulaire du Général de Gaulle}.
\newblock Armand Colin, Presses de la Fondation nationale des sciences
  politiques, 1969.

\bibitem{deerwester:1990}
S.~Deerwester, S.~T. Dumais, G.~W. Furnas, T.~K. Landauer, and R.~Harshman.
\newblock Indexing by latent semantic analysis.
\newblock {\em American Society For Information Science}, 41:391--407, 1990.

\bibitem{el-beze:2005}
M.~El-Bèze, J.-M. Torres-Moreno, and F.~Béchet.
\newblock Peut-on rendre automatiquement à césar ce qui lui appartient ?
  {A}pplication au jeu du {C}hirand-{M}itterrac.
\newblock In {\em Proc. of TALN 2005 - DEFT'05}, volume~2, pages 125--134,
  2005.

\bibitem{freund:1997}
Y.~Freund and R.E. Schapire.
\newblock A decision-theoretic generalization of online learning and an
  application to boosting.
\newblock {\em Journal of Computer and System Sciences}, 55:119--139, 1997.

\bibitem{Minimerror}
M.B. Gordon and D.~Berchier.
\newblock Minimerror: A perceptron learning rule that finds the optimal
  weights.
\newblock In Michel Verleysen, editor, {\em ESANN}, pages 105--110, Brussels,
  1993. D facto.

\bibitem{hofmann:1999}
T.~Hofmann.
\newblock Probabilistic latent semantic analysis.
\newblock In {\em Proc. of 2nd Annual ACM Conference on Research and
  Development in Information Retrieval, Berkeley, California}, pages 50--57,
  1999.

\bibitem{illouz:2000}
G.~Illouz, B.~Habert, S.~Fleury, H.~Folch, S.~Heiden, P.~Lafon, and
  S.~Prévost.
\newblock Profilage de textes : cadre de travail et expérience.
\newblock In {\em JADT 2000, Journées Int. d'Analyse Statistiques des Données
  Textuelles, Lausanne}, pages 163--170, 2000.

\bibitem{jalam:2002}
R.~Jalam and J.-H. Chauchat.
\newblock Pourquoi les n-grammes permettent de classer des textes ? recherche
  de mots-clefs pertinents à l'aide des n-grammes caractéristiques.
\newblock In {\em JADT 2002, Journées Int. d'Analyse Statistiques des Données
  Textuelles, St-Malo}, pages 13--15, 2002.

\bibitem{katz:1987}
S.~M. Katz.
\newblock Estimation of probabilities from sparse data for the language model
  component of a speech recognizer.
\newblock {\em IEEE Transactions on Acoustics, Speech and Signal Processing},
  35:400--401, 1987.

\bibitem{labbe:l990}
D.~Labbé.
\newblock {\em Le vocabulaire de {F}rançois {M}itterrand}.
\newblock Presses de la Fondation Nationale des Sciences Politiques, mars 1990,
  Paris, 1990.

\bibitem{manning:2000}
C.~D. Manning and H.~Schütze.
\newblock {\em Foundations of Statistical Natural Language Processing}.
\newblock The MIT Press, 2000.

\bibitem{rigouste:2005}
L.~Rigouste, C.~Olivier, and Y.~François.
\newblock Modèle de mélange multi-thématique pour la fouille de textes.
\newblock In {\em Proc. of TALN 2005 - Atelier DEFT'05}, volume~2, pages
  193--202, 2005.

\bibitem{sahami:1999}
M.~Sahami.
\newblock {\em Using Machine Learning to Improve Information Access}.
\newblock Phd thesis, Computer Science Department, Stanford University., 1999.

\bibitem{schapire:2000}
R.E. Schapire and Y.~Singer.
\newblock {B}oos{T}exter: {A} boosting-based system for text categorization.
\newblock {\em Machine Learning}, 39(2/3):135--168, 2000.

\bibitem{soboro:2004}
I.~Soboro.
\newblock Overview of the {TREC} 2004 {N}ovelty {T}rack.
\newblock In E.~M. Voorhees and Lori~P. Buckland, editors, {\em Proceedings of
  TREC'04}, USA, 2004. NIST Special Publication: SP.

\bibitem{spriet:1996}
T.~Spriet, F.~Béchet, M.~El-Bèze, C.~De Loupy, and L.~Khouri.
\newblock Traitement automatique des mots inconnus.
\newblock In {\em Proc. of TALN 96}, pages 170--179., Marseille France, 22-24
  mai, 1996.

\bibitem{spriet:1998}
T.~Spriet and M.~El-Bèze.
\newblock Introduction of {R}ules into a {S}tochastic {A}pproach for {L}anguage
  {M}odelling.
\newblock {\em Computational Models of Speech Pattern Processing},
  169:350--355, 1998.

\bibitem{torres-moreno:2002}
J.-M. Torres-Moreno, J.C. Aguilar, and M.B. Gordon.
\newblock Finding the number minimum of errors in {N}-dimensional parity
  problem with a linear perceptron.
\newblock {\em Neural Processing Letters}, 1:201--210, 2002.

\bibitem{Monoplane}
J.-M. Torres~Moreno and M.~Gordon.
\newblock An evolutive architecture coupled with optimal perceptron learning.
\newblock In Michel Verleysen, editor, {\em ESANN}, pages 365--370, Brussels,
  1995. D facto.

\bibitem{eurowordnet:1998}
P.~Vossen.
\newblock {\em EuroWordNet: A Multilingual Database with Lexical Semantic
  Networks}.
\newblock Kluwer Academic publishers, 1998.

\end{thebibliography}

\section*{Annexe}

Nous souhaitons illustrer le fonctionnement du modèle de cohérence interne en rapportant ici une étude sur le discours 520 du corpus de test, dont toutes les phrases appartiennent à la classe $C$ (Chirac). Le tableau \ref{tab:Annexe} montre des extraits des phrases de ce discours, ainsi que leurs probabilités $p(C)$ d'appartenance à la classe $C$ (probabilité $p(M) = 1-p(C)$) calculées par adaptation et celles $p_I$ obtenues avec la cohérence interne du discours (cf. section \ref{sec:cohérence}), puis adaptation. L'étiquette associée est aussi indiquée. En gras, nous affichons les mots pleins ({\bfseries Afrique, africains, Congo, développ(er/ment)},...) des phrases étiquetées $C$. En souligné les mots pleins communs (\underline{effort}, \underline{institutions}) aux deux classes, et en petites majuscules les mots pleins (\textsc{accord}) présents uniquement à l'intérieur du bloc $M$. Les adjectifs, adverbes, pronoms, conjonctions de subordination ainsi que les noms propres trop fréquents dans le corpus DEFT'05 (tels que {\slshape France, français(e), Paris, international, pays,}...) ont été supprimés pour ne pas fausser les calcu\-ls.
\begin{figure}[ht]
\begin{center}
 \includegraphics[width=10cm]{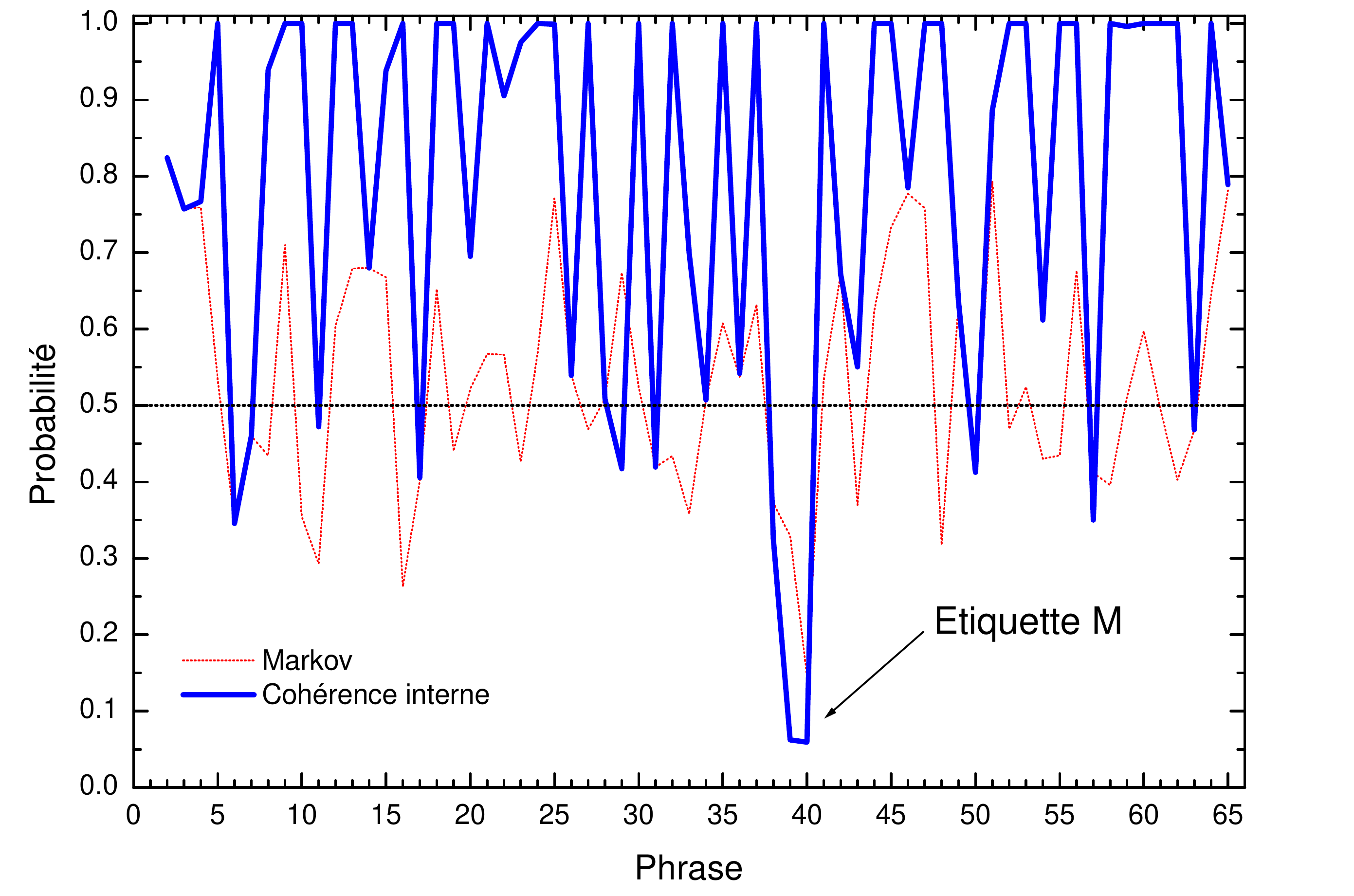}
 \caption{Probabilités $p(C)$ de Markov (ligne pointillée) et de la cohérence interne puis adaptation   
         (trait gras) du discours 520 (dont tous les segments appartiennent à la classe C).} 
 \label{fig:probabilites}
\end{center}
\end{figure}
Les phrases 39 et 40 (en italiques) méritent d'être analysées en détail. Leurs probabilités d'appartenance à la classe $M$, calculées uniquement avec l'adaptation seule, étaient déjà très élevées : 0,671 ($p(C)=0,329$) et 0,853 ($p(C)=0,147$) respectivement (et sont d'ailleurs les plus élevées de tout ce discours), donc difficiles à renverser. Elles faisaient partie d'un gros bloc (phrases 31-40) étiqueté $M$ par Markov. Après le calcul de la cohérence, les phrases 39 et 40 seront encore étiquetées comme $M$, avec des valeurs 0,91 ($p(C)=0,09$) et 0,89 ($p(C)=0,11$) respectivement. Cependant, la méthode de la cohérence interne a fait basculer vers la classe $C$ sept phrases (31-38) du bloc original, étiquetées $M$ à tort dans un premier temps. Ce renversement n'est en rien négligeable comme en atteste l'augmentation du Fscore due au mécanisme de la cohérence. La figure \ref{fig:probabilites} montre les probabilités $p(C)$ en fonction du numéro de la phrase du discours. En pointillé, nous affichons les probabilités de l'adaptation seule et en gras continu celles de la cohérence interne, puis adaptation. On voit que, sauf pour les deux phrases en question, toutes les autres ont été réarrangées de façon à avoir des probabilités $p(C)$ bien au-dessus de leurs valeurs précédentes (plusieurs phrases ont maintenant des probabilités $p(C)=1$). La rupture de cohérence interne du discours reflète ici plutôt un changement thématique : tout le discours 520 s'inscrit fortement dans une vision politique centrée sur l'Afrique, alors que les phrases 39 et 40 parlent soudain de politique internationale dans un sens beaucoup plus large ({\slshape New York, Washington, institutions internationales}). Ceci illustre la tendance (toujours difficile à modéliser) qu'a parfois un locuteur d'introduire subitement des changements (quasi aléatoires) dans son discours. Cet honneur revient dans le cas présent, à Jacques Chirac.

\begin{table}
 \begin{center}
   \tabcolsep = .5\tabcolsep
   \begin{tabular}{ccc}
   \hline\hline
      Discours 520 (Test) & \multicolumn{2}{c}{Probabilités $C$}\\
      Phrase & $p$ & $p_I$ \\
\hline
 \begin{minipage}{10cm}
 \footnotesize{
    ...13 
    ... remercier tout particulièrement l'ensemble d'entre vous qui, civils, militaires, religieux, coopérants, industriels, professionnels libéraux, artisans, commerçants, etc... sont la France, la France au \textbf{Congo}.
    }
  \end{minipage} & \footnotesize{0,68 C} & \footnotesize{1,00 C}\\
 \begin{minipage}{10cm}
 \footnotesize{ 
    ...21 
    un peu partout, et notamment au \textbf{Congo}, l'état de droit, la démocratie, est en train de prendre racine et de se \textbf{développer}.
    }
  \end{minipage} & \footnotesize{0,57 C} & \footnotesize{1,00 C}\\
 \begin{minipage}{10cm}
 \footnotesize{ 
    ...24 De la même façon, on voit les \underline{efforts} considérables qui sont engagés en \textbf{Afrique}, en \textbf{Afrique} francophone, au \textbf{Congo}, pour libéraliser l'économie, sortir des \underline{structures} paralysantes, qui, longtemps, ont caractérisé beaucoup de pays \textbf{africains} pour donner une nouvelle... 
    }
  \end{minipage} & \footnotesize{0,57 C} & \footnotesize{1,00 C}\\
 \begin{minipage}{10cm}
 \footnotesize{ 
    ...27 On connaît encore des problèmes, des crises, ici ou là, pas loin du \textbf{Congo}, dans cette \textbf{Afrique}.
    }
  \end{minipage} & \footnotesize{0,47 C} & \footnotesize{1,00 C}\\
 \begin{minipage}{10cm}
 \footnotesize{ 
    ...29 On voit des \underline{efforts} de coordination régionale qui sont engagés et qui permettront une plus grande synergie de \underline{l'effort} économique et ... 
    }
  \end{minipage} & \footnotesize{0,67 C} & \footnotesize{0,42 C}\\
 \begin{minipage}{10cm}
 \footnotesize{ 
    30 Je le disais à Franceville tout à l'heure, il y a cinq ans, il y avait peine 20 pays qui, en \textbf{Afrique}, avaient une croissance positive... 
    }
  \end{minipage} & \footnotesize{0,53 C} & \footnotesize{1,00 C}\\
 \begin{minipage}{10cm}
 \footnotesize{ 
    31 Il y en a 41 aujourd'hui en quelques années.
    }    
  \end{minipage} & \footnotesize{0,42 M} & \footnotesize{0,51 C}\\
 \begin{minipage}{10cm}
 \footnotesize{ 
    32 Celles ou ceux qui sont \textbf{afro} pessimistes sont nombreux, notamment dans le reste du monde, (certains parce qu'ils sont tout simplement découragés, certains parce qu'ils sont ignorants ou veulent l'être de ce qui se passe ici, certains parce qu'ils veulent en fait se désengager de l'aide que le monde industrialisé doit au titre de la solidarité à l'\textbf{Afrique} en
 \textbf{développement}).
    }
  \end{minipage} & \footnotesize{0,43 M} & \footnotesize{1,00 C}\\
 \begin{minipage}{10cm}
  \footnotesize{
    33 Il y a une espèce de culte d'\textbf{afro} pessimisme, il faut aujourd'hui comprendre qu'il n'y a plus aucune raison de \textbf{développer} ce sentiment.
    }
  \end{minipage} & \footnotesize{0,36 M} & \footnotesize{0,78 C}\\
\begin{minipage}{10cm}
 \footnotesize{
    34 
    peut aujourd'hui, raisonnablement, justement, être \textbf{afro} optimiste.
    }
  \end{minipage} & \footnotesize{0,51 M} & \footnotesize{0,58 C}\\
\begin{minipage}{10cm}
 \footnotesize{
    35 La croissance du \textbf{Congo} est de 6\%.
    }
  \end{minipage} & \footnotesize{0,61 M} & \footnotesize{1,00 C}\\
\begin{minipage}{10cm}
 \footnotesize{
    36 J'imagine la satisfaction des Français s'ils faisaient la même performance, nous n'aurions plus aucun problème.
    }
  \end{minipage} & \footnotesize{0,54 M} & \footnotesize{0,99 C}\\
\begin{minipage}{10cm}
 \footnotesize{
    37 Il faut encourager les \textbf{Africains}.
    }
  \end{minipage} & \footnotesize{0,63 M} & \footnotesize{1,00 C}\\
\begin{minipage}{10cm}
 \footnotesize{
    38 
    pas assez les \underline{efforts} considérables, parfois avec des maladresses dues souvent à l'inexpérience, qu'ils ont fait pour \textbf{redresser} la \underline{situation}.
    }
  \end{minipage} & \footnotesize{0,37 M} & \footnotesize{0,39 C}\\
   \hline 
\begin{minipage}{10cm}
 \footnotesize{
    39 \slshape{Il y a quelques années, seuls quelques pays avaient un \underline{\textsc{accord}} avec les \underline{institutions} internationales, aujourd'hui presque tous sont dans... 
    }
    }
  \end{minipage} & \footnotesize{\textbf{\slshape{0,33 M}}} & \footnotesize{\textbf{\slshape{0,09 M}}}\\
\begin{minipage}{10cm}
 \footnotesize{
    40 \slshape{Je sais bien qu'il est de bon ton, je l'ai fait moi-même souvent, de critiquer des \underline{institutions} internationales qui, depuis New York ou Washington, depuis les bureaux climatisés et qui sont là-bas à partir des ordinateurs qui s'y trouvent, imposent des règles, non seulement extrêmement difficiles à accepter dans les pays qui doivent faire un \underline{effort} d'ajustement \underline{structurel}, mais de plus, le font souvent dans des termes qui ne sont même pas compris, ici, là où leurs règles ... 
    }
    }
  \end{minipage} & \footnotesize{\textbf{\slshape{0,15 M}}} & \footnotesize{\textbf{\slshape{0,11 M}}}\\
   \hline
\begin{minipage}{10cm}
 \footnotesize{
    41 Mais il faut dire aussi, que le temps passant, il y a une amélioration sensible de l'approche, de la vision portée par ces \underline{institutions} sur l'\textbf{Afrique}.
    }
  \end{minipage} & \footnotesize{0,54 C} & \footnotesize{1,00 C}\\
\begin{minipage}{10cm}
 \footnotesize{
    ...43 Alors des progrès sont accomplis, il faut naturellement tout faire pour les \textbf{développer}.
    }
  \end{minipage} & \footnotesize{0,37 M} & \footnotesize{0,64 C}\\
\begin{minipage}{10cm}
 \footnotesize{
    44 La France, vous le savez, est très attachée à sa politique \textbf{africaine}.
    }
  \end{minipage} & \footnotesize{0,63 M} & \footnotesize{1,00 C}\\
\begin{minipage}{10cm}
 \footnotesize{
    45 Elle l'est bien sûr en raison des liens très anciens qui nous unissent, l'\textbf{Afrique} ou une partie de l'\textbf{Afrique} et nous-mêmes.
    }
  \end{minipage}  & \footnotesize{0,73 M} & \footnotesize{1,00 C}\\
\begin{minipage}{10cm}
 \footnotesize{
    ...47 C'est en \textbf{Afrique} qu'elle a puisé l'énergie, le courage, la détermination, le sang qui nous a permis de \textbf{redresser} notre \underline{situation} si compromise.
    }
  \end{minipage}  & \footnotesize{0,76 M} & \footnotesize{1,00 C}\\
\begin{minipage}{10cm}
 \footnotesize{
    ...57 Il y a eu, c'est, vrai, une décennie mauvaise, certains ont dit une décennie perdue, et bien, nous nous en sommes sortis, et donc maintenant nous devons faire un \underline{effort}.
    }
  \end{minipage}  & \footnotesize{0,41 M} & \footnotesize{0,41 C}\\
\begin{minipage}{10cm}
 \footnotesize{
    ...64 Et bien, je voudrais vous donner ce soir, c'est mon dernier mot, un message de confiance et d'espoir, d'encouragement aussi, et vous dire que la France est fière d'avoir ses meilleurs enfants sur cette terre d'\textbf{Afrique}, au \textbf{Congo} ou ailleurs, et vous dire que votre rôle est capital pour cette terre et aussi pour les valeurs, les valeurs morales qui sont... 
    }
  \end{minipage}  & \footnotesize{0,65 M} & \footnotesize{1,00 C}\\
   \hline
   \end{tabular}
 \caption{Exemple de découpage de la cohérence interne. Discours 520 du Test.}
 \label{tab:Annexe}
 \end{center}
\end{table}



\end{document}